%% file: main.tex
\pdfoutput=1

\documentclass[11pt]{article}

\usepackage[final]{acl}
\usepackage{times}
\usepackage{latexsym}

\usepackage[T1]{fontenc}

\usepackage[utf8]{inputenc}

\usepackage{microtype}
\usepackage{inconsolata}
\usepackage{hyperref}
\usepackage{url}
\usepackage{booktabs}
\usepackage{lineno}
\usepackage{graphicx}
\usepackage{enumitem}
\usepackage{comment}
\usepackage{caption}
\usepackage{xcolor}
\usepackage[edges]{forest}
\usepackage{tabularx}
\usepackage{multirow}
\usepackage{booktabs}
\usepackage{makecell}
\usepackage[table]{xcolor}
\usepackage{pifont}
\usepackage{nicematrix}

\definecolor{darkblue}{rgb}{0, 0, 0.5}
\hypersetup{colorlinks=true, citecolor=darkblue, linkcolor=darkblue, urlcolor=darkblue}

\definecolor{devcolor}{HTML}{FEE7E5}   
\definecolor{cogcolor}{HTML}{FFF7DB}   
\definecolor{behavcolor}{HTML}{FFE7CC} 
\definecolor{soccolor}{HTML}{E2F1E1}   
\definecolor{perscolor}{HTML}{E5EEFA}  
\definecolor{lingcolor}{HTML}{F8E5FF}  

\definecolor{psychdefcolor}{HTML}{034545}

\newcommand{\gary}[1]{\textcolor{blue}{gary: #1}}
\newcommand{\zack}[1]{\textcolor{brown}{zack: #1}}

\newcommand{\psychTheory}[1]{\textbf{\textit{#1}}}
\newcommand{\psychDef}[1]{\textbf{\textit{\textcolor{psychdefcolor}{#1}}}}

\title{A Review of Incorporating Psychological Theories in LLMs}

\author{Zizhou Liu$^{\ast\spadesuit}$,~~~~ 
Ziwei Gong$^{\ast\spadesuit}$,~~~~ 
Lin Ai$^{\ast\spadesuit}$,~~~~ 
Zheng Hui$^{\ast\dagger}$,~~~~ 
Run Chen$^{\ast\spadesuit}$,\\ 
\textbf{Colin Wayne Leach$^\diamond$},~~~~  
\textbf{Michelle R. Greene$^\diamond$},~~~~ 
\textbf{Julia Hirschberg$^\spadesuit$}\\
\small{$^\ast$Equal contributions.}\\ 
  $^\spadesuit$Department of Computer Science, Columbia University, \\
  $^\diamond$Department of Psychology, Barnard College,\\
  $^\dagger$The Language Technology Lab, University of Cambridge
\\
\small{
    \texttt{\{sara.ziweigong, lin.ai, runchen,julia\}@cs.columbia.edu}
  }
  \\
\small{
    \texttt{\{cleach, mgreene\}@barnard.edu,\{zl2889, zh2483\}@columbia.edu}
  }
}

\begin{document}

\maketitle

\begin{abstract}
Psychological insights have long shaped pivotal NLP breakthroughs, from attention mechanisms to reinforcement learning and social modeling. As Large Language Models (LLMs) develop, there is a rising consensus that psychology is essential for capturing human-like cognition, behavior, and interaction.
This paper reviews how psychological theories can inform and enhance stages of LLM development. Our review integrates insights from six subfields of psychology, including cognitive, developmental, behavioral, social, personality psychology, and psycholinguistics. With stage-wise analysis, we highlight current trends and gaps in how psychological theories are applied. By examining both cross-domain connections and points of tension, we aim to bridge disciplinary divides and promote more thoughtful integration of psychology into NLP research.

\end{abstract}


\section{Introduction}
\input{cog_dev_papers/intro}




\section{Preprocessing}
\begin{figure*}[ht]
    \centering
    \resizebox{0.97\textwidth}{!}{  
    \input{graph/overview}
    }
    \captionsetup{font=small}
    \caption{Our structured survey of how psychological theories apply across the main stages of LLM development. Colors indicate six distinct psychology areas: red for \colorbox{devcolor}{\psychTheory{Developmental Psychology}}; orange for \colorbox{behavcolor}{\psychTheory{Behavioral Psychology}}; yellow for \colorbox{cogcolor}{\psychTheory{Cognitive Psychology}}; green for \colorbox{soccolor}{\psychTheory{Social Psychology}}; blue for \colorbox{perscolor}{\psychTheory{Personality Psychology}}; purple for \colorbox{lingcolor}{\psychTheory{Psycholinguistics}}.}
    \label{fig:taxonomy}
    \vspace{-3mm}
\end{figure*}
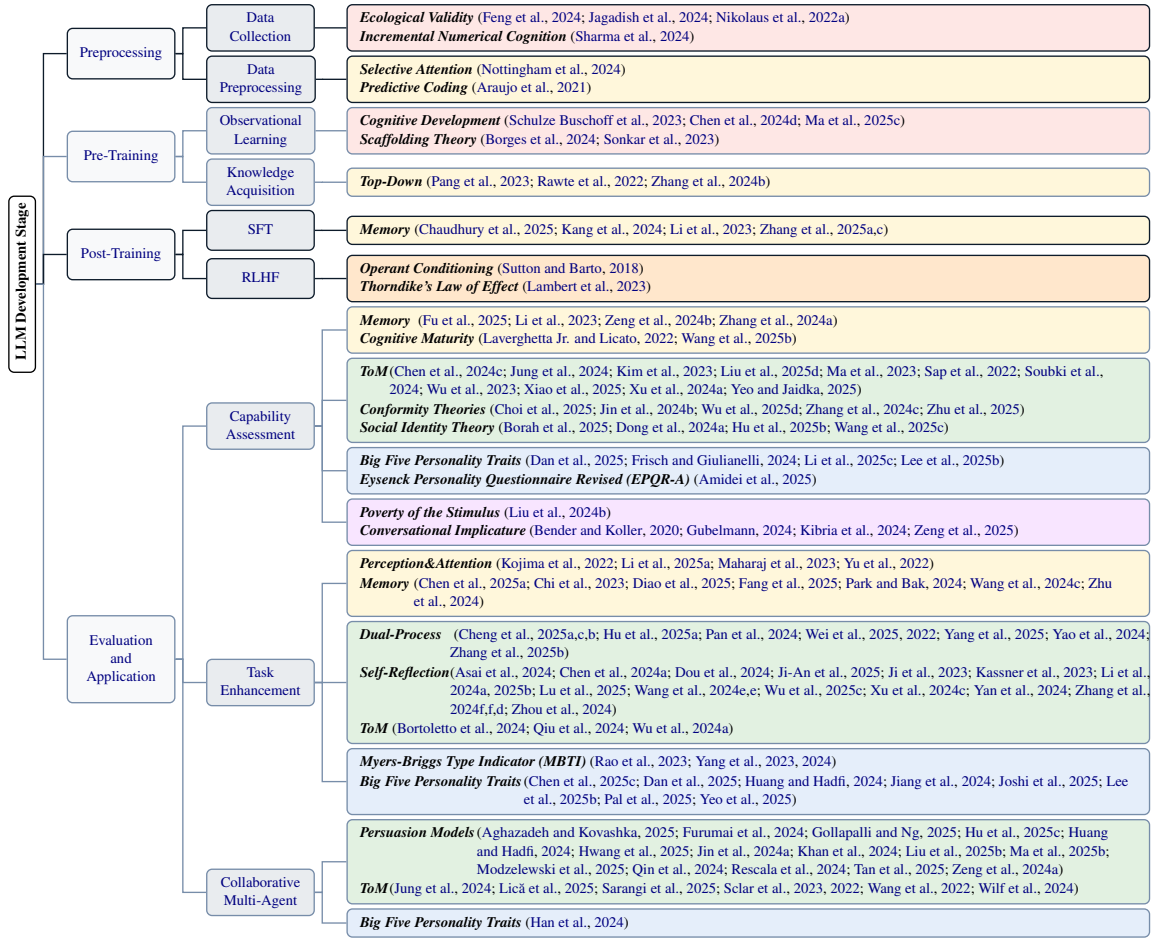
\label{sec:preprocessing}
\input{preprocessing/data}

\section{Pre-Training}
\label{sec:pretraining}

\input{pre_training/pre_training}

\section{Post-Training and Alignment}
\label{sec:posttraining}
\input{post_training/post_training}

\section{Evaluation and Application}
\label{sec:application}
Psychology offers tools for both assessing and enhancing model behavior in the evaluation and application stage. We review three key areas of challenges where psychology can inform LLM development: \textbf{(1)} evaluating emergent capabilities such as reasoning, \textbf{(2)} improving task performance in domains involving human cognition, and \textbf{(3)} designing socially aware, multi-agent systems.

\input{application/application}

\section{Trends and Discussion}
\label{sec:discussion}
\input{dicussion/discussion}

\section{Conclusions}
We systematically review how psychology can ground LLM innovation in both past and future across six key subfields. We examine how psychological theories inform each stage of LLM development, revealing both meaningful connections across domains and critical points of tension, which we explore through discussion to help bridge interdisciplinary gaps. We hope this survey sparks reflection, and inspires future work to continue integrating psychological perspectives into NLP in meaningful and impactful ways.



\section*{Limitations}
Our review
primarily focuses on literature within NLP, particularly in how personality is modeled, evaluated, and leveraged in LLMs. As a result, we do not extensively cover research from psychology and cognitive sciences that might offer deeper theoretical insights into human-like behaviors in AI. This limitation may exclude valuable methodologies or perspectives that could enhance personality evaluation frameworks for LLMs. We encourage future surveys to integrate findings from psychology and linguistics to bridge theoretical foundations with computational approaches, fostering a more comprehensive understanding of personality in AI systems. Some perspectives (e.g., embodied cognition, constructivism) cut across our categories and appear sporadically in current LLM work; we note these as cross-cutting threads and encourage future surveys to map them more fully across stages."

While our survey advocates for a deeper integration of psychology into LLM design, we also caution against the ethical risks posed by overuse or misapplication of psychological principles. A concrete example is \psychTheory{operant conditioning} \citep{skinner1957behavior}, which describes how behavior can be shaped by consequences. Applied to LLMs, these mechanisms can be beneficial in contexts like language learning or motivation. For instance, through timely, gratifying feedback to reinforce engagement. However, reinforcement schedules such as variable ratio or interval rewards may unintentionally condition users to engage compulsively, raising the risk of manipulative design. This presents a key ethical limitation: distinguishing between genuinely supportive interactions and those that encourage excessive use is inherently difficult. To address this, we emphasize the need for transparent disclosure of reinforcement mechanisms and the establishment of clear ethical guidelines by professional communities. These safeguards are essential to ensure that psychological insights enhance user well-being without enabling exploitative practices.

We also recognize longstanding critiques of psychology, for example, its emphasis on individual-level constructs, WEIRD sampling (overreliance on participants from Western, Educated, Industrialized, Rich, and Democratic societies in behavioral and social science research), and under-representation of social/structural forces. Our survey foregrounds psychology because many current LLM papers explicitly reference it; however, insights from linguistics and sociology are equally important for understanding language as a fundamentally social phenomenon. Future surveys that braid these traditions more tightly would provide a more complete view.

An important adjacent question is what LLMs do psychologically to users (e.g., persuasion, behavior change, support). We flag this as a crucial direction for future surveys that integrate psychology with HCI and communication research to evaluate user-level outcomes alongside model-level capabilities.


\section*{Acknowledgments}
 This work is in part supported by the funds provided by the National Science Foundation and by DoD OUSD (R\&E) under Cooperative Agreement PHY-2229929 (The NSF AI Institute for Artificial and Natural Intelligence). The views, opinions and/or findings expressed are those of the authors and should not be interpreted as representing the official views or policies of the National Science Foundation or the U.S. Government.

\bibliography{anthology, cog_dev_papers/cog_dev, behavioral/behavioral, social/social, language_papers/language, personality/personality, extra_references}

\clearpage
\appendix
\input{appendix/psychology_theories}

\input{appendix/keywords_validation}

\input{appendix/rlhf}
\input{appendix/dialgue}
\input{appendix/debates}

\section{Scope of Psychology in the Paper}

By “psychology,” we refer to theories about cognition, development, behavior, emotion, social interaction, and individual differences—traditionally situated within psychological science. These domains overlap with linguistics (form/meaning/usage), HCI (interaction), and sociology (social structure), yet they maintain distinct emphases. For instance, linguistics often focuses on formal structures and rule systems, such as phonological constraints or morphological productivity, which can be studied independently of human cognition. In contrast, psycholinguistics investigates how such structures are acquired, processed, and used by the mind, for example, how humans manage ambiguity via Conversational Implicature or generalize from limited input as posited by the Poverty of the Stimulus hypothesis. Similarly, sociology addresses macro-level forces such as institutional inequality or class-based language stratification, while social psychology operates at the meso-level, exploring how individual behavior is shaped by group dynamics and identity, exemplified by theories such as Conformity and Social Identity Theory.

Our taxonomy is a pragmatic organizing device, co-developed with psychology collaborators; it is not a claim of strict disciplinary separation. Where theories cross boundaries, we note this and treat psychology as one complementary lens.

\end{document}

%% file: cog_dev_papers/intro.tex
As Large Language Models (LLMs) grow in scale and complexity, the Natural Language Processing (NLP) community increasingly sees psychology as key to capturing human-like cognition, behavior, and interaction \citep{qu2024promoting,lewisartificial}. Psychology, grounded in empirically validated and computationally adaptable frameworks \citep{sartori2023language,ong2024gpt}, offer principled tools for addressing persistent LLM challenges such as reasoning fidelity, context retention, and user interaction. Early intersections of psychology and NLP predate modern LLMs. Classic systems such as ELIZA \cite{10.1145/365153.365168} framed conversations around simple pattern matching, while early cognitive and psycholinguistic work influenced representation and processing assumptions. Today’s LLMs revive these threads at a new scale, but psychology is best viewed as one productive lens among others, complementary to linguistics, HCI, and sociology, rather than an inevitable or singular foundation. Reflecting these strengths, psychological insights have driven NLP advances, including the cognitive inspirations of attention mechanisms, formative reinforcement learning approaches, and social modeling for agents.

Despite extensive multidisciplinary efforts, a holistic review systematically integrating psychology across the LLM lifecycle remains missing. Most surveys and position papers remain fragmented, typically falling into three broad categories: \textbf{(1)} Some investigate how LLMs can empower traditional psychology or cognitive science research, for instance by modeling human reasoning and behavior at scale  \citep{abdurahman2024perils,demszky2023using,ong2024gpt,ke2024exploringfrontiersllmspsychological}. \textbf{(2}) Others approach LLMs as subjects of psychological analysis, aiming to adapt or extend psychological theory, such as personality or cognition frameworks, to interpret and evaluate model behavior \citep{brito-etal-2025-modeling,li2024quantifyingaipsychologypsychometrics,2023arXiv230313988H,huang2024on,doi:10.1177/17456916231214460}. \textbf{(3)} Finally, a third group leverages a single or limited set of psychological constructs to enhance model alignment or multi-agent frameworks -- improving system reliability, social interaction, and trustworthiness \citep{dong2024survey,liu2023trustworthy}. This includes research on social influence for AI safety \citep{zeng-etal-2024-johnny}, moral reasoning in legal tasks \citep{almeida2024exploring}, and partial integrations of social or developmental psychology \citep{sartori2023language,serapiogarcía2025personalitytraitslargelanguage,zhang-etal-2024-exploring}. However, no existing work provides a unified map of how diverse psychological sub-areas can be harnessed, from data through application. Our survey fills this gap by offering a stage-wise view of how psychology can strengthen LLM capabilities and alignment across the entire lifecycle.

To address this gap, we present a structured review that situates psychological theories from six major areas across the entire LLM development pipeline. The contributions of our survey\footnote{We survey 227 papers from major *CL venues, plus COLING, NeurIPS, ICML, ICLR, and influential arXiv preprints. Appendix \ref{appendix:keywords} details paper selection process.} are twofold:
\textbf{(1)} We systematically review psychological theories applied in key stages of LLM development, identifying gaps and inconsistencies.
\textbf{(2)} We highlight under-explored concepts  alongside critical issues and debates at the intersection of psychology and NLP.
Collectively, these contributions demonstrate how integrating diverse psychological frameworks can strengthen LLM design, enhance alignment, and broaden the practical and ethical impact
of modern NLP systems.

As shown in Figure~\ref{fig:taxonomy}, the remainder of this paper illustrates how cognitive, developmental, behavioral, social, psycholinguistic, and personality theories integrate into four key stages of LLM development: preprocessing (Section~\ref{sec:preprocessing}), pre-training (Section~\ref{sec:pretraining}), post-training (Section~\ref{sec:posttraining}), and evaluation and application (Section~\ref{sec:application}). Finally, Section~\ref{sec:discussion} discusses three central questions: \emph{(i) How does current LLM development leverage psychological theories? (ii) Which untapped psychological insights could advance LLM development? (iii) What debates loom at the intersection of NLP and psychology?}

%% file: graph/overview.tex
\definecolor{stagecolor}{HTML}{f8f9fa}
\definecolor{subseccolor}{HTML}{e9ecef}
\definecolor{preprocesscolor}{HTML}{0d1b2a}
\definecolor{pretraincolor}{HTML}{778da9}
\definecolor{posttraincolor}{HTML}{0d1b2a}
\definecolor{appcolor}{HTML}{778da9}

\begin{forest}
forked edges,
for tree={
    grow'=east,
    parent anchor=east,
    child anchor=west,
    anchor=base west,
    rounded corners,
    edge+={line width=1pt, draw=black},
    draw,
    font=\normalsize,
    content format={\strut\forestoption{content}},
    l sep=0.8cm,
    s sep=0.1 cm,
    fit=band,
    calign=center,
    where level=0{tier=0}{},
    where level=1{tier=1}{},
    where level=2{tier=2}{},
    where level=3{tier=3}{},
}
[ {\rotatebox{90}{\textbf{LLM Development Stage}}}, fill=white, inner sep=5pt, line width=1pt
  [{\protect\hyperref[sec:preprocessing]{Preprocessing}}, fill=stagecolor, inner ysep = 12, minimum width=80pt, edge={draw=preprocesscolor}, line width=1pt, draw=preprocesscolor
    [{\begin{tabular}{c}
      \hyperref[subsec:data_collection]{Data} \\ 
      \hyperref[subsec:data_collection]{Collection}
    \end{tabular}}, fill=subseccolor, minimum width=80pt, edge={draw=preprocesscolor}, line width=1pt, draw=preprocesscolor
      [
      {\begin{minipage}{590pt}
        \begin{tabular}{l}
          \psychTheory{Ecological Validity}~\citep{feng-etal-2024-child, jagadish2024injecting, nikolaus-etal-2022-learning}
        \end{tabular}
        \\
        \begin{tabular}{l}
          \psychTheory{Incremental Numerical Cognition}~\citep{sharma-etal-2024-laying}
        \end{tabular}
      \end{minipage}}, fill=devcolor, edge={draw=preprocesscolor}, line width=1pt, draw=preprocesscolor
      ]
    ]
    [{\begin{tabular}{c}
      \hyperref[subsec:data_preprocessing]{Data} \\ 
      \hyperref[subsec:data_preprocessing]{Preprocessing}
    \end{tabular}}, fill=subseccolor, minimum width=80pt, edge={draw=preprocesscolor}, line width=1pt, draw=preprocesscolor
      [
      {\begin{minipage}{590pt}
        \begin{tabular}{l}
           \psychTheory{Selective Attention}~\citep{nottingham-etal-2024-selective}
        \end{tabular}
        \\
        \begin{tabular}{l}
          \psychTheory{Predictive Coding}~\citep{araujo-etal-2021-augmenting}
        \end{tabular}
      \end{minipage}}, fill=cogcolor, edge={draw=preprocesscolor}, line width=1pt, draw=preprocesscolor
      ]
    ]
  ]
  [{\protect\hyperref[sec:pretraining]{Pre-Training}}, fill=stagecolor, inner ysep = 12, minimum width=80pt, edge={draw=pretraincolor}, line width=1pt, draw=pretraincolor
    [
    {\begin{tabular}{c}
      \hyperref[subsec:observational_learning]{Observational} \\ 
      \hyperref[subsec:observational_learning]{Learning}
    \end{tabular}}, fill=subseccolor, minimum width=80pt, edge={draw=pretraincolor}, line width=1pt, draw=pretraincolor
      [
      {\begin{minipage}{590pt}
        \begin{tabular}{l}
           \psychTheory{Cognitive Development}~\citep{buschoff2023acquisition, chen-etal-2024-temporal, ma-etal-2025-babysit}
        \end{tabular}
        \\
        \begin{tabular}{l}
          \psychTheory{Scaffolding Theory}~\citep{borges-etal-2024-teach, sonkar-etal-2023-class}
        \end{tabular}
      \end{minipage}}, fill=devcolor, edge={draw=pretraincolor}, line width=1pt, draw=pretraincolor
      ]
    ]
    [{\begin{tabular}{c}
      \hyperref[subsec:knowledge_acquisition]{Knowledge} \\ 
      \hyperref[subsec:knowledge_acquisition]{Acquisition}
    \end{tabular}}, fill=subseccolor, minimum width=80pt, edge={draw=pretraincolor}, line width=1pt, draw=pretraincolor
      [
      {\begin{minipage}{590pt}
        \begin{tabular}{l}
          \psychTheory{Top-Down}~\citep{pang-etal-2023-long, rawte2022tdlr, zhang-etal-2024-holistic}
        \end{tabular}
      \end{minipage}}, fill=cogcolor, edge={draw=pretraincolor}, line width=1pt, draw=pretraincolor
      ]
    ]
  ]
  [{\protect\hyperref[sec:posttraining]{Post-Training}}, fill=stagecolor, inner ysep = 12, minimum width=80pt, edge={draw=posttraincolor}, line width=1pt, draw=posttraincolor
    [{\hyperref[subsec:sft]{SFT}}, fill=subseccolor, minimum width=80pt, inner ysep=8, edge={draw=posttraincolor}, line width=1pt, draw=posttraincolor
      [
      {\begin{minipage}{590pt}
        \begin{tabular}{l}
          \psychTheory{Memory}~\citep{chaudhury-etal-2025-epman, kang2024think, li-etal-2023-large, zhang-etal-2025-growing, zhang-etal-2025-prime}
        \end{tabular}
      \end{minipage}}, fill=cogcolor, edge={draw=posttraincolor}, line width=1pt, draw=posttraincolor
      ]
    ]
    [{\hyperref[subsec:rlhf]{RLHF}}, fill=subseccolor, minimum width=80pt, inner ysep=8, edge={draw=posttraincolor}, line width=1pt, draw=posttraincolor
      [
      {\begin{minipage}{590pt}
        \begin{tabular}{l}
          \psychTheory{Operant Conditioning}~\citep{sutton2018reinforcement} \\
          \psychTheory{Thorndike’s Law of Effect}~\citep{lambert2023historyrisksreinforcementlearning}\\
      \end{tabular}
      \end{minipage}}, fill=behavcolor, edge={draw=posttraincolor}, line width=1pt, draw=posttraincolor
      ]
    ]
  ]
  [{\protect\hyperref[sec:application]{
  \begin{tabular}{c}
       \textnormal{Evaluation}  \\ 
       \textnormal{and} \\
       \textnormal{Application}
  \end{tabular}}}, fill=stagecolor, inner ysep = 12, minimum width=80pt, edge={draw=appcolor}, line width=1pt, draw=appcolor
    [{\begin{tabular}{c}
      \hyperref[subsec:benchmarks]{Capability} \\ 
      \hyperref[subsec:benchmarks]{Assessment}
    \end{tabular}}, fill=subseccolor, minimum width=80pt, edge={draw=appcolor}, line width=1pt, draw=appcolor
      [
      {\begin{minipage}{590pt}
        \begin{tabular}{l}
          {\psychTheory{Memory}} ~\citep{fu-etal-2025-exclusion, li-etal-2023-large, zeng-etal-2024-memorize, zhang-etal-2024-working}
\end{tabular}
\\
\begin{tabular}{l}
  \psychTheory{Cognitive Maturity}~\citep{laverghetta-jr-licato-2022-developmental, wang-etal-2025-coglm}
        \end{tabular}
        \end{minipage}}, fill=cogcolor, edge={draw=appcolor}, line width=1pt, draw=appcolor
      ]
      [
      {\begin{minipage}{590pt}
        \begin{tabular}{lp{530pt}}
          \makebox[10pt][l]{\psychTheory{ToM}} & \citep{chen-etal-2024-tombench, jung-etal-2024-perceptions, kim-etal-2023-fantom, liu-etal-2025-tactfultom, ma-etal-2023-towards-holistic, sap-etal-2022-neural, soubki-etal-2024-views, wu-etal-2023-hi, xiao-etal-2025-towards, xu-etal-2024-opentom, yeo-jaidka-2025-beyond}
        \end{tabular}
        \\
        \begin{tabular}{l}
          \psychTheory{Conformity Theories}~\citep{choi-etal-2025-empirical, jin-etal-2024-agentreview, wu-etal-2025-exploring-choice, zhang-etal-2024-exploring,zhu-etal-2025-conformity}
        \end{tabular}
        \\
        \begin{tabular}{l}
          \psychTheory{Social Identity Theory}~\citep{borah-etal-2025-mind, dong2024persona, hu2025generative, wang-etal-2025-evaluating-cognitive}
        \end{tabular}
      \end{minipage}}, fill=soccolor, edge={draw=appcolor}, line width=1pt, draw=appcolor
      ]
      [
      {\begin{minipage}{590pt}
        \begin{tabular}{l}
          \psychTheory{Big Five Personality Traits}~\citep{dan-etal-2025-p, frisch-giulianelli-2024-llm, li-etal-2025-big5, lee-etal-2025-llms}
        \end{tabular}
        \\
        \begin{tabular}{l}
          \psychTheory{Eysenck Personality Questionnaire Revised (EPQR-A)} \citep{amidei-etal-2025-exploring}
        \end{tabular}
      \end{minipage}}, fill=perscolor, edge={draw=appcolor}, line width=1pt, draw=appcolor
      ]
      [
      {\begin{minipage}{590pt}
        \begin{tabular}{l}
          \psychTheory{Poverty of the Stimulus} \citep{liu2024zhoblimpsystematicassessmentlanguage}
        \end{tabular}
        \\
        \begin{tabular}{lp{390pt}}
          \makebox[114pt][l]
          {\psychTheory{Conversational Implicature}}&  \citep{bender-koller-2020-climbing, gubelmann-2024-pragmatic, kibria-etal-2024-functional, zeng-etal-2025-converging}
        \end{tabular}
      \end{minipage}}, fill=lingcolor, edge={draw=appcolor}, line width=1pt, draw=appcolor
      ]
    ]
    [{\begin{tabular}{c}
      \hyperref[subsec:task_enhancement]{Task} \\ 
      \hyperref[subsec:task_enhancement]{Enhancement}
    \end{tabular}}, fill=subseccolor, minimum width=80pt, edge={draw=appcolor}, line width=1pt, draw=appcolor
      [
      {\begin{minipage}{590pt}
        \begin{tabular}{l}
          \psychTheory{Perception\&Attention}~\citep{10.5555/3600270.3601883, li2025cam,maharaj-etal-2023-eyes,yu-etal-2022-relation}
        \end{tabular}
    \\
    \begin{tabular}{p{28pt}p{520pt}}
      {\psychTheory{Memory}} & \citep{chen-etal-2025-improving, chi-etal-2023-transformer, diao-etal-2025-temporal, fang2025hippocampallike, park2024memoria, wang-etal-2024-symbolic, zhu-etal-2024-beyond}
    \end{tabular}
\end{minipage}}, fill=cogcolor, edge={draw=appcolor}, line width=1pt, draw=appcolor
      ]
      [
      {\begin{minipage}{590pt}
        \begin{tabular}{lp{520pt}}
          \makebox[55pt][l]{\psychTheory{Dual-Process}} & ~\citep{cheng-etal-2025-dualrag, cheng-etal-2025-think, cheng2025incentivizing, hu-etal-2025-stitchllm, pan-etal-2024-dynathink, wei-etal-2025-mecot, wei2022chain, yang-etal-2025-llm2, yao2024tree, zhang-etal-2025-leveraging}
        \end{tabular}
        \\
        \begin{tabular}{lp{520pt}}
          \makebox[55pt][l]{\psychTheory{Self-Reflection}} & \citep{asai2024selfrag, chen-etal-2024-dual, dou-etal-2024-rest, ji-an2025language, ji-etal-2023-towards, kassner-etal-2023-language, li-etal-2024-hindsight, li-etal-2025-adaptive, lu2025auditing, wang-etal-2024-taste, wang-etal-2024-taste, wu-etal-2025-seeval, xu-etal-2024-sayself, yan-etal-2024-mirror, zhang-etal-2024-learn, zhang-etal-2024-learn, zhang-etal-2024-self-contrast, zhou2024metacognitive}
        \end{tabular}
        \\
        \begin{tabular}{l}
          \psychTheory{ToM}~\citep{bortoletto-etal-2024-limits, qiu-etal-2024-minddial, wu-etal-2024-coke}
        \end{tabular}
      \end{minipage}}, fill=soccolor, edge={draw=appcolor}, line width=1pt, draw=appcolor
      ]
      [
      {\begin{minipage}{590pt}
        \begin{tabular}{l}
          \psychTheory{Myers-Briggs Type Indicator (MBTI)}~\citep{rao-etal-2023-chatgpt, yang-etal-2023-psycot, yang-etal-2024-psychogat}
        \end{tabular}
        \\
        \begin{tabular}{lp{450pt}}
          \makebox[110pt][l]{\psychTheory{Big Five Personality Traits}} & \citep{chen2025synthempathyscalableempathycorpus, dan-etal-2025-p, huang-hadfi-2024-personality, jiang-etal-2024-personallm, joshi-etal-2025-improving, lee-etal-2025-llms, pal-etal-2025-beyond, yeo-etal-2025-pado}
        \end{tabular}
      \end{minipage}}, fill=perscolor, edge={draw=appcolor}, line width=1pt, draw=appcolor
      ]
    ]
    [{\begin{tabular}{c}
      \hyperref[subsec:multi_agent]{Collaborative} \\ 
      \hyperref[subsec:multi_agent]{Multi-Agent}
    \end{tabular}}, fill=subseccolor, minimum width=80pt, edge={draw=appcolor}, line width=1pt, draw=appcolor
      [
      {\begin{minipage}{590pt}
        \begin{tabular}{lp{470pt}}
          \makebox[75pt][l]{\psychTheory{Persuasion Models}} & \citep{aghazadeh-kovashka-2025-face, furumai-etal-2024-zero, gollapalli-ng-2025-pirsuader, hu-etal-2025-debate, huang-hadfi-2024-personality, hwang-etal-2025-trick, jin-etal-2024-persuading, 10.5555/3692070.3693020, liu-etal-2025-synthetic, ma-etal-2025-enhancing, modzelewski-etal-2025-pcot, qin-etal-2024-beyond, rescala-etal-2024-language, tan-etal-2025-persuasion, zeng-etal-2024-johnny}
        \end{tabular}
        \\
        \begin{tabular}{lp{550pt}}
          \makebox[10pt][l]{\psychTheory{ToM}} & \citep{jung-etal-2024-perceptions, lica2025mindforge, sarangi-etal-2025-decompose, sclar-etal-2023-minding, sclar2022symmetric, wang2021tomc, wilf-etal-2024-think}
        \end{tabular}
      \end{minipage}}, fill=soccolor, edge={draw=appcolor}, line width=1pt, draw=appcolor
      ]
      [
      {\begin{minipage}{590pt}
      \begin{tabular}{l}
        \psychTheory{Big Five Personality Traits} \citep{han-etal-2024-psydial}
      \end{tabular}
      \end{minipage}}, fill=perscolor, edge={draw=appcolor}, line width=1pt, draw=appcolor
      ]
    ]
  ]
]
\end{forest}

%% file: preprocessing/data.tex

We begin the stage-by-stage analysis of LLM development with preprocessing, the foundation that shapes downstream capabilities. Psychology provides valuable frameworks for understanding how humans acquire and filter information, underscoring the need for realistic, developmentally informed datasets and effective filtering strategies in data collection and processing.

\input{preprocessing/data_collection}

\input{preprocessing/data_preprocessing}

%% file: preprocessing/data_collection.tex
\textbf{Data Construction~} Recent evidence shows that LLMs can align with human brain responses under biologically plausible training conditions \citep{hosseini2024artificial}, despite LLMs typically requiring orders of magnitude more training data than human. This supports the application of \psychTheory{ecological validity} \citep{schmuckler2001ecological} that \psychDef{emphasizes real-world data to mimic cognitive development}. To reflect children's language acquisition processes, \citet{jagadish2024injecting} selects linguistically diverse environments, \citet{feng-etal-2024-child} utilizes child-directed speech, while \citet{peppapig} collects child cartoon. In parallel, \psychTheory{incremental numerical understanding} \citep{Piaget1953-PIATCC-4} that \psychDef{views numerical concepts as gradually acquired through exposure} is applied to sequential data collection with mathematically coherent numeric anchors \citep{sharma-etal-2024-laying}. Lastly, \citet{reuben-etal-2025-assessment} provides a systematic framework to reformulate psychological questionnaires for LLM assessment. 
\label{subsec:data_collection}

%% file: preprocessing/data_preprocessing.tex
\textbf{Data Preprocessing~} Data preprocessing inspired by cognitive psychology involves refining data to enhance informational coherence prior to training. \psychTheory{Selective attention} \citep{treisman1969selective_attention}, \psychDef{prioritizing salient information while filtering out irrelevant stimuli}, was implemented to develop a preprocessing model that filters irrelevant data \citep{nottingham-etal-2024-selective}. Meanwhile, \psychTheory{predictive coding} proposing \psychDef{anticipatory processing based on prior knowledge} \citep{rao1999predictive}, was leveraged by \citet{araujo-etal-2021-augmenting} to enable anticipation of subsequent content, improving semantic coherence through expectation-driven processing. Lastly, drawing insights from \psychTheory{knowledge acquisition of children}, \citet{ficarra-etal-2025-distributional} redefines lexical knowledge in data to capture distributional information based on target word.

\label{subsec:data_preprocessing}


%% file: pre_training/pre_training.tex

Building on the foundations established during preprocessing, pre-training mirrors human cognitive development, where linguistic and reasoning abilities emerge through exposure to stimuli. 
This section explores how psychology inform observational learning and knowledge acquisition in LLMs.


\textbf{Observational Learning }
\label{subsec:observational_learning}
\input{cog_dev_papers/Observational_Learning_and_Self-Supervision}

\textbf{Knowledge Acquisition}
\label{subsec:knowledge_acquisition}
\input{cog_dev_papers/World_Knowledge_Acquisition}

%% file: cog_dev_papers/Observational_Learning_and_Self-Supervision.tex
\psychTheory{Incremental cognitive development} \citep{piaget1976piaget}, which posits \psychDef{children acquire knowledge through sequential tasks}, informs how LLMs can master nuanced concepts with explicit structured exposure. It is used as a motivating analogy for curriculum and sequence effects \citep{bengio2009curriculum} (i.e., how staged exposure can scaffold capability growth) rather than to claim a theoretical identity with next-token prediction. This principle manifests in \citet{buschoff2023acquisition}'s gradually expanding pre-training tasks, \citet{chen-etal-2024-temporal}'s contradictory historical tasks and \citet{ma-etal-2025-babysit}'s trial-and-demonstration framework. In contrast, self-supervised objectives align more closely with \psychTheory{use-based views of meaning}, resonating with Wittgenstein’s claim that meaning emerges through use \citep{Strawson1954-STRWL}.
Additionally, \psychTheory{scaffolding theory} \citep{park2009adaptive}, which \psychDef{emphasizes gradually challenging interactions}, informs maintaining coherent learning trajectories through \citet{borges-etal-2024-teach}'s structured feedback loops and \citet{sonkar-etal-2023-class}'s dynamic task complexity.

%% file: cog_dev_papers/World_Knowledge_Acquisition.tex
Semantic coherence during pre-training draw insights from \psychTheory{top-down and bottom-up perception} \citep{gregory_topdown_bottomup}, which \psychDef{frames cognition as interaction between conceptual frameworks and detailed data}. Top-down processing is leveraged to prioritize semantic processing before syntactic details \citep{rawte2022tdlr} and to generate test cases \citep{zhang-etal-2024-holistic}. Meanwhile, to enhance perception modeling, \citet{pang-etal-2023-long} fuses bottom-up encoding with top-down corrections, and \citet{nikolaus-fourtassi-2021-modeling} models production-based learning.
Introducing \psychTheory{working memory theory} \citep{baddeley1974working} that proposes \psychDef{a short-term system for temporarily holding information}, \citet{mita-etal-2025-developmentally} simulates critical period dynamics with growing memory capacity to enhance performance.


%% file: post_training/post_training.tex
With foundational knowledge acquired in pre-training, post-training refines LLMs from general proficiency to task-specific behavior. We explore how psychology guide post-training for context-aware, interpretable, and human-aligned outcomes.

\input{post_training/sft}

\input{post_training/rlhf}

%% file: post_training/sft.tex
\textbf{Supervised Fine-Tuning (SFT)~} In SFT, works that draw on psychological insights focus on retaining and learn contextual information. Building on \psychTheory{working memory theory}, \citet{kang2024think} adds a working memory module to retain short-term information, while \citet{li-etal-2023-large} dynamically balances memory with contexts to improve robustness. Drawing from \psychTheory{episodic memory}, \psychDef{the ability to retrieve specific experiences with contexts}\citep{tulving1972episodic}, \citet{zhang-etal-2025-growing} enable LLMs to learn from episodic experiences for improved planning, \citet{zhang-etal-2025-prime} integrates episodic memory with semantic memory for LLM personalization, while \citet{chaudhury-etal-2025-epman} introduce episodic attention for processing long contexts.
\label{subsec:sft}

%% file: post_training/rlhf.tex
\textbf{Reinforcement Learning from Human Feedback (RLHF)} 
\label{subsec:rlhf}
A classic behavioral theory, the \psychTheory{Operant Conditioning theory} posits that \psychDef{behaviors are systematically strengthened or weakened by the consequences (rewards or punishments) that immediately follow them} \citep{skinner1957behavior,thorndike1898animal} principles of reinforcement learning align closely with this psychological framework, particularly in the post-training phase of LLM development, where RLHF explicitly operationalizes \psychTheory{Operant Conditioning theory} to align model behaviors with human values and preferences. Through repeated feedback, the model gradually adapts to favor outputs that yield higher reward signals---a process akin, in a loose analogy, to \psychTheory{Thorndike’s Law of Effect}, which describes how \psychDef{behaviors followed by satisfying outcomes tend to recur}. While the underlying mechanism is driven by reward optimization algorithms rather than psychological intent, the conceptual resemblance highlights how reinforcement strategies can shape model outputs \citep{lambert2023historyrisksreinforcementlearning}.
During RLHF, the model generates responses, and a learned reward function $R(x)$ assigns scores to outputs $x$, guiding subsequent policy updates. For instance, \citet{ouyang2022training} train InstructGPT using Proximal Policy Optimization \citep{schulman2017proximal}, rewarding responses preferred by humans and penalizing less desirable ones. Foundational frameworks \citep{NIPS2017_d5e2c0ad, stiennon2022learningsummarizehumanfeedback,sutton2018reinforcement}explicitly translating human judgments into reward signals, operationalizing the insights of \psychTheory{Operant Conditioning}. More recent work incorporates human cognitive biases \citep{siththaranjan2024distributional, wadi-fredette-2025-monte} and personalizes reward functions for individual values \citep{poddar2024personalizing} and awareness \citep{liu-etal-2025-cogdual}. These developments illustrate how \psychTheory{Operant Conditioning} remains central to aligning LLMs with nuanced human values. While our survey focuses on psychological dimensions, a technical overview of RLHF methods is provided in Appendix~\ref{appendix:rlhf}.

%% file: application/application.tex

\input{application/benchmarks}

\input{application/task_enhancement}
\input{application/multiagent}

%% file: application/benchmarks.tex
\subsection{Benchmarks and Capability Assessment}
\label{subsec:benchmarks}

Evaluating LLMs with psychologically grounded metrics offers a deeper window into their real-world viability. By mapping classic theories onto benchmarks that probe model responses under diverse, human-like scenarios, researchers can move beyond surface-level performance measures, revealing emergent model behavior and illuminating strengths, blind spots, and opportunities to refine LLM training and alignment practices.

\subsubsection{Social Reasoning and Intelligence}
Social intelligence is vital for LLMs that navigate human contexts, enabling the interpretation of implicit cues, adaptation to social norms, and authentic interaction -- defining advanced AI beyond mere text prediction. As LLMs increasingly mediate communication, their grasp of social dynamics becomes pivotal for both efficacy and safety.

Notably, \psychTheory{Theory of Mind (ToM)} offers a framework for evaluating \psychDef{how individuals understand and attribute mental states -- such as beliefs, desires, and intentions -- to others.}
By measuring LLMs' capacity to reason beliefs, researchers can assess core social intelligence. Recent benchmarks probe different facets of \psychTheory{ToM}: \citet{chen-etal-2024-tombench, xu-etal-2024-opentom} provide relatively comprehensive coverage, while others test ToM on  higher-order reasoning \citep{wu-etal-2023-hi}, interactions \citep{kim-etal-2023-fantom}, white lies \citep{liu-etal-2025-tactfultom}, cognitive appraisal \citep{yeo-jaidka-2025-beyond}, and temporally evolving mental states \citep{xiao-etal-2025-towards}. Extending the efforts to spoken dialogues, \citet{soubki-etal-2024-views} reveal lingering gaps between LLM and human performance. Surveys \citep{chen-etal-2025-theory, ma-etal-2023-towards-holistic, sap-etal-2022-neural, soubki-rambow-2025-machine} consolidate methods and underscore the challenges of robust \psychTheory{ToM}-based evaluations.

Beyond individual cognition, social influence theories like \psychTheory{Conformity Theories} \citep{asch2016effects}, capture \psychDef{how group pressure shapes individual judgments}. Recent work tests LLM-based agents’ collaboration and bias dynamics under these principles \citep{choi-etal-2025-empirical, jin-etal-2024-agentreview, wu-etal-2025-exploring-choice, zhang-etal-2024-exploring,zhu-etal-2025-conformity}, bridging individual and group-level cognition.

Emotion is another pillar of social intelligence. \psychTheory{Ekman’s Basic Emotion Theory} \citep{ekman1992there} identifies \psychDef{six universal emotions}, often used as labels, while \psychTheory{Dimensional Models} like the \textit{Circumplex Model} conceptualize emotions along valence and arousal \citep{gong-etal-2024-mapping, morrill-etal-2024-social}. LLMs advance on emotion recognition, benefiting dialogue and sentiment tasks \citep{sabour-etal-2024-emobench, wu2024silentlettersamplifyingllms, wu-etal-2024-multimodal,zhang-etal-2024-sentiment}.

These efforts collectively demonstrate both progress and limitations in LLMs' social cognition, establishing benchmarks against which future developments can be measured.

\subsubsection{Language Proficiency}
Recent work adopts psycholinguistic assessments, originally designed for humans, to test LLMs’ language proficiency.
These experiments probe a wide range of linguistic domains: morphology \citep{anh-etal-2024-morphology}, syntax \citep{amouyal-etal-2025-lm,an2025hierarchical, hale-stanojevic-2024-llms,kandra-etal-2025-llms,li-hao-2025-eras, liu2024zhoblimpsystematicassessmentlanguage}, phonology \citep{goriely-buttery-2025-babylms, jang-etal-2025-p}, semantics \citep{ bavaresco-fernandez-2025-experiential,duan-etal-2025-unveiling, hayashi-2025-evaluating} and their combinations \citep{ma-etal-2025-pragmatics, miaschi-etal-2024-evaluating, zhou-etal-2025-linguistic}. 

Although LLMs exhibit comparable performance to human speakers on many psycholinguistic tasks, the underlying processing mechanism they rely on may seem different from humans \citep{lee-etal-2024-psycholinguistic, pedrotti-etal-2025-humans}. Human language acquisition is often characterized by the \psychTheory{Poverty of the Stimulus}, \psychDef{children acquire complex grammar from relatively little input} \citep{Chomsky_1980}, whereas LLMs typically require developmentally implausible amounts of linguistic data to learn morphological rules. 
On the other hand, some evidence suggests that the learning patterns of LLMs mirror aspects of human language acquisition \citep{liu2024zhoblimpsystematicassessmentlanguage, zhou-etal-2025-context}.

Several studies have explored the pragmatic abilities of LLMs, motivated by the close link between language and broader cognitive functions in humans. \psychTheory{\citet{Grice1975-GRILAC-6}’s Theory of conversational implicature} posits that \psychDef{utterance interpretation depends on both literal content and surrounding context}.
Researchers \citep{bender-koller-2020-climbing, gubelmann-2024-pragmatic} have contrasting perspectives on LLMs with respect to the \citet{HARNAD1990335}'s \psychTheory{Symbol Grounding Problem}, i.e. \psychDef{linguistic symbols must be grounded in sensorimotor interactions to be meaningful}.
Failures of LLMs in pragmatic and semantic tasks \citep{he-etal-2025-large-language, junker-etal-2025-multimodal,kibria-etal-2024-functional, zeng-etal-2025-converging, ma-etal-2025-pragmatics}, as well as their neuron patterns \citep{he-etal-2025-large-language,wu-etal-2024-rethinking}, point to limitations beyond pure linguistic knowledge, which potentially parallel human higher-level cognitive processes. 


\subsubsection{Memory and Cognitive Evaluation}
Assessing memory and cognition is crucial given LLMs' limited capacity and risk of catastrophic forgetting. \psychTheory{Memory} is measured on parametric knowledge \citep{li-etal-2023-large}, n-back tasks \citep{zhang-etal-2024-working}, capacity \citep{timkey-linzen-2023-language} and \psychTheory{cognitive load} \citep{fu-etal-2025-exclusion, xu-etal-2024-cognitive, zeng-etal-2024-memorize}. 
Meanwhile, cognitive development is assessed through \psychTheory{cognitive maturity} \citep{laverghetta-jr-licato-2022-developmental, wang-etal-2025-coglm}, word acquisition \citep{chang-bergen-2022-word}, \psychTheory{subjective similarity} \citep{malloy-etal-2024-leveraging}, reasoning strategies \citep{mondorf-plank-2024-comparing,yuan-etal-2023-beneath, ying-etal-2024-intuitive}, \psychTheory{zone of proximal} development \citep{cui-sachan-2025-investigating} and \psychTheory{perception} \citep{jung-etal-2024-perceptions, zhou2025scientists}.

\subsubsection{Personality Capability}

\input{personality/personalityCapibility}
\subsubsection{Bias and Ethics Evaluation}
Evaluating biases and ethical risks is crucial for responsible AI that avoids reinforcing harmful social patterns. As LLMs increasingly shape public discourse, thorough assessments are essential to prevent discriminatory outputs and promote equitable benefits across diverse communities. Recent work tests LLMs on gender \citep{oba-etal-2024-contextual, zhao-etal-2024-comparative}, broader social biases \citep{abolghasemi-etal-2025-evaluation, huang-etal-2025-fact, lee-etal-2023-kosbi, muti-etal-2025-r, nozza-etal-2022-pipelines, shin-etal-2024-ask, zhao-etal-2025-explicit}, toxic content \citep{gehman-etal-2020-realtoxicityprompts, huang-etal-2025-intrinsic, hui2024can, luong-etal-2024-realistic}, and harmful stereotypes \citep{grigoreva-etal-2024-rubia, huang-xiong-2024-cbbq, hui2024toxicraft, shrawgi-etal-2024-uncovering}, establishing benchmarks across cultures and languages. Evidence also suggests that LLMs replicate social identity biases, mirroring human tendencies toward ingroup favoritism and outgroup hostility \citep{borah-etal-2025-mind, dong2024persona, hu2025generative, wang-etal-2025-evaluating-cognitive} -- patterns central to \psychTheory{social identity theory}, which posits that \psychDef{group membership shapes self-concept and intergroup behavior} 
\citep{tajfel1979integrative}.

%% file: personality/personalityCapibility.tex

Personality consistency examines how stably LLMs maintain traits across contexts. \citet{frisch-giulianelli-2024-llm} show LLMs with asymmetric profiles vary in \psychTheory{Big Five} traits, while \citet{amidei-etal-2025-exploring} find language switching alters GPT-4o’s \psychTheory{Eysenck Personality Questionnaire Revised} traits, underscoring challenges in perserving consistency. 
Parallel research examines how LLMs display and control personality traits. \citet{jiang-etal-2024-personallm} show LLMs express distinct traits labeled by human evaluators. \citet{10.1007/978-981-97-9434-8_19} reveals difficulties in alignment for \psychTheory{Neuroticism}, \psychTheory{Extraversion} and \psychTheory{Agreeableness}. \citet{han-etal-2025-value, lee-etal-2025-llms, li-etal-2025-big5, li-etal-2025-decoding-llm} assess and improve consistency through alignment with psychometrical training data, while \citet{hu-collier-2024-quantifying} find persona-based prompting improves annotation accuracy. 

%% file: application/task_enhancement.tex
\subsection{Task Performance Enhancement}
\label{subsec:task_enhancement}

Building on the benchmarks, we review how psychological insights are used to improve LLMs performance on complex reasoning and enrich dialogue, which illustrate how psychology improves capabilities and alignment across applications.

\subsubsection{Reasoning Enhancement}



LLMs often struggle with complex reasoning: social inference \citep{liu2024exploring}, logical errors \citep{mckenna-etal-2023-sources, turpin2023language}, hallucinations \citep{ai-etal-2024-enhancing, huang2025survey}, and multi-step planning \citep{wang2024q}. Researchers address these issues by implementing analogous cognitive mechanisms. For instance, \psychTheory{Dual-process theory}, a social cognition framework, \psychDef{distinguishes between fast (System 1) and slow (System 2) reasoning} \citep{kahneman2011thinking}, offers a blueprint for LLM improvement. Chain-of-thought prompting \citep{wei2022chain} operationalizes System 2 via intermediate steps, while DynaThink \citep{pan-etal-2024-dynathink} dynamically selects rapid or thorough inference. Tree of Thoughts \citep{yao2024tree} further explores multiple reasoning paths concurrently. \citet{yang-etal-2025-llm2} combine separate verifier as System 2. More recent applications includes hallucination mitigation \citep{cheng-etal-2025-think}, real-time human-AI collaboration \citep{zhang-etal-2025-leveraging}, multi-hop QA \citep{cheng-etal-2025-dualrag}, emotion consistency \citep{wei-etal-2025-mecot}, decoder-level LLMs merging \citep{hu-etal-2025-stitchllm} and cost-efficient RL framework \citep{cheng2025incentivizing}.

Similarly, \psychTheory{Self-reflection and Meta-cognition}, \psychDef{introspection focused on the self-concept} \citep{flavell1979metacognition, Phillips2020}, has guided LLM enhancements in hallucination mitigation \citep{ji-etal-2023-towards,lu2025auditing}, translation \citep{chen-etal-2024-dual, wang-etal-2024-taste}, tool use \citep{li-etal-2025-adaptive}, question-answering \citep{kassner-etal-2023-language, li-etal-2024-hindsight, zhang-etal-2024-learn}, retrieval-augmented-generation(RAG) \citep{asai2024selfrag, zhou2024metacognitive} and math reasoning \citep{zhang-etal-2024-learn}. Approaches include iterative self-assessment \citep{ji-etal-2023-towards, wu-etal-2025-seeval, yan-etal-2024-mirror}, task decomposition \citep{wang-etal-2024-taste, zhang-etal-2024-self-contrast}, self-training \citep{dou-etal-2024-rest}, in context learning \citep{ji-an2025language}, and confidence-tuned reward functions \citep{xu-etal-2024-sayself}.
Moreover, \psychTheory{ToM} adaptations boost LLMs’ interpersonal reasoning, aiding missing knowledge \citep{bortoletto-etal-2024-limits}, common ground alignment \citep{qiu-etal-2024-minddial}, and cognitive modeling \citep{wu-etal-2024-coke}.

Beyond social reasoning, \psychTheory{perception, attention, and memory} support coherence and retrieval. \citet{10.5555/3600270.3601883} uses “think step by step” prompts for \psychTheory{top-down} reasoning. \citet{li2025cam} employed \psychTheory{Piaget's constructivist theory} on a systematic memory system for reading comprehension. \citet{chen-etal-2025-improving, maharaj-etal-2023-eyes,yu-etal-2022-relation} leverages \psychTheory{selective attention} and \psychTheory{working memory} to detect hallucinations and extract relation. \citet{zhu-etal-2024-beyond} employs recitation for retrieval, and \citet{park2024memoria} introduce short/long-term memory modules. \citet{chi-etal-2023-transformer, diao-etal-2025-temporal, wang-etal-2024-symbolic} improve reasoning via \psychTheory{symbolic, adaptive and working memory} structures. Lastly, \psychTheory{hippocampal indexing theory} \citep{teyler1986hippocampal}, \psychDef{viewing the hippocampus as a pointer to neocortical memory}, informs multi-step reasoning with external knowledge \citep{fang2025hippocampallike, gutierrez2024hipporag} and counterfactual reasoning \citep{miao-etal-2024-episodic}. 

\subsubsection{Dialogue Understanding and Generation}
\label{subsubsec:dialogue_understanding}
In dialogue understanding, personality psychology aids trait-based inferences from user interactions. NLP research has explored dynamic ways to measure personality beyond structured tests. The \psychTheory{Myers–Briggs Type Indicator (MBTI)}, \psychDef{a self-report questionnaire that makes pseudo-scientific claims to categorize individuals into 16 distinct personality types}, remains popular \citep{rao-etal-2023-chatgpt, yang-etal-2023-psycot}, while PsychoGAT \citep{yang-etal-2024-psychogat} gamifies \psychTheory{MBTI}, and \textit{PADO} \citep{yeo-etal-2025-pado} adopts a \psychTheory{Big Five}-based multi-agent approach. Beyond assessments, traits guide dialogue generation: \citet{huang-hadfi-2024-personality} show higher agreeability improves negotiation, while \citet{cheng-etal-2023-marked} reveal social and racial biases in persona creation, raising representational concerns.

Dialogue generation research further incorporates personality to improve coherence, empathy, and consistency. \citet{dan-etal-2025-p} enforced trait-consistent generation via \psychTheory{Big Five} trait specialization loss. \citet{joshi-etal-2025-improving} augmented personas with rationales generated by \psychTheory{Big Five}-guided scaffolds. \citet{chen2025synthempathyscalableempathycorpus, pal-etal-2025-beyond} leveraged Reddit-based journal entries to model \psychTheory{Big Five} traits in large-scale dialogue datasets. \psychTheory{Big Five}-aligned agents also improve on text based games \citep{lim-etal-2025-persona}, decision making under risk \citep{hartley-etal-2025-personality} and code generation \citep{guo-etal-2025-personality}. Other efforts improve persona consistency without referencing explicit psychological theory \citep{takayama-etal-2025-persona, wu-etal-2025-traits}. Similarly, personality is used to improve truthfulness, consistency, and context-aware generation, as further detailed in Appendix \ref{appendix:persona_dialogue}. 
These approaches support personality alignment but lack grounding in deeper psychological theory.

%% file: application/multiagent.tex
\subsection{Collaborative, Multi-Agent Frameworks}
\label{subsec:multi_agent}

Beyond task-specific capabilities, the surge in multi-agent LLM frameworks reflects a growing emphasis on collaborative decision-making, where modeling social dynamics is crucial. Social and personality psychology theories offer insights to design agent interaction, negotiation, and consensus, guiding more socially intelligent LLM systems.

\paragraph{Social Influence} \psychTheory{Persuasion models} \citep{petty2012communication} illustrate \psychDef{how central/peripheral routes shape attitudes in collaborative settings}. Leveraging this, \citet{gollapalli-ng-2025-pirsuader} merges persuasive dialog acts with RL, \citet{modzelewski-etal-2025-pcot} infuses persuasion knowledge into CoT, \citet{furumai-etal-2024-zero} combines LLM strategies and retrieval, \citet{jin-etal-2024-persuading, qin-etal-2024-beyond} emphasize credibility-aware generation, and \citet{hwang-etal-2025-trick, tan-etal-2025-persuasion,zeng-etal-2024-johnny} uncovers LLMs' vulnerabilities. Multi-agent research simulates personality-driven \citep{huang-hadfi-2024-personality,hu-etal-2025-debate,liu-etal-2025-synthetic} and cultural-driven negotiation \citep{aghazadeh-kovashka-2025-face, ma-etal-2025-enhancing}, boosts truthfulness via structured debates \citep{10.5555/3692070.3693020}, and curates argument-strength datasets \citep{rescala-etal-2024-language}.

\paragraph{Social Cognition}\psychTheory{ToM} complements social influence by enabling agents to grasp others’ mental states. Some integrate perspective-taking \citep{lica2025mindforge}, belief tracking \citep{sclar-etal-2023-minding} and coordination \citep{wang2021tomc, sclar2022symmetric}, while others refine \psychTheory{ToM} via task decomposition and recursive simulation \citep{jung-etal-2024-perceptions, sarangi-etal-2025-decompose, wilf-etal-2024-think}.

\paragraph{Role-Play and Multi-Agent Simulation}
Recent work on persona-driven LLM agents focuses on simulating diverse perspectives, persona alignment, and socially intelligent interactions. \citet{han-etal-2024-psydial} introduces \psychTheory{Big Five}-based extraversion, \citet{castricato-etal-2025-persona} presents 1,586 synthetic personas, and \citet{wu-etal-2025-raiden} releases a benchmark with 40K multi-turn dialogues. Agents also model opinion dynamics \citep{wang-etal-2025-decoding} and evaluate social intelligence \citep{chen-etal-2024-socialbench}, with RoleLLM \citep{wang-etal-2024-rolellm}, Character100 \citep{wang-etal-2024-characteristic}, and persona-aware graph transformers \citep{mahajan-shaikh-2024-persona} further supporting multi-party simulations. Lastly, \citet{kumarage2025personalized} and \citet{chen2026detectingmentalmanipulationspeech} simulate social engineering attacks with LLM and Audio LM agents of varied traits, highlighting how psychological traits shape user vulnerability and the need for more robust defenses against personalized manipulation.

%% file: dicussion/discussion.tex

\subsection{How Does Current LLM Development Harness Psychological Theories?}
\input{dicussion/discussion_trends}

\subsection{What Untapped Psychological Insights Could Advance LLM Development?}

\input{dicussion/missing_theory}

\subsection{What Debates Loom at the NLP– Psychology Intersection, and Where Next?}

\input{dicussion/discussion_debate}

%% file: dicussion/discussion_trends.tex
From the review from Section \ref{sec:preprocessing}-\ref{sec:application}, 
we observe psychological theories have been incorporated into LLM development in stage-specific ways, with uneven coverage across theoretical domains. Figure~\ref{fig:taxonomy} maps this integration across stages.

In early stages, preprocessing and pretraining, \textbf{developmental psychology} is often referenced. Its emphasis on staged knowledge acquisition aligns with curriculum learning and progressive data exposure, mirroring human developmental trajectories.
In post-training, especially RLHF, \textbf{behavioral psychology} ideas are most prominent. Conditioning, reinforcement schedules, and reward design are commonly used to guide model alignment with human preferences.
In evaluation and application, theories from \textbf{social psychology, personality psychology} and \textbf{psycholinguistics} are commonly cited, reflecting a focus on interaction patterns, user modeling, and linguistic variation -- areas traditionally explored within these sub-fields. Their prominence in later stages aligns with their emphasis on human-centered communication.
\textbf{Cognitive psychology} appears across all stages, particularly in modeling internal mechanisms such as reasoning, memory, and attention. Its breadth makes it a foundational influence, though less visible in agentic interaction settings.

The observed unevenness in integration reflects, perhaps a gap, but more probably a functional alignment -- some domains are naturally better suited for certain stages of LLM development.
Meanwhile, these trends expose under-explored opportunities, motivating the RQs that follow.

%% file: dicussion/missing_theory.tex

Although psychological theory is increasingly applied in LLM research, its use remains simplified and uneven. 
As shown in Tables \ref{tab:psych_theories_dev_cog_beh}, \ref{tab:psych_theories_soc}, and \ref{tab:psych_theories_per_ling}, many theories are under-utilized despite their potential to improve model behavior and interpretability. Below, we outline theories in four key areas that deserve greater attention in future LLM research.

\textbf{Social psychology} remains underutilized in areas like \psychTheory{group dynamics} and \psychTheory{self and identity}, limiting personalization, adaptability, and inclusivity. Prompting LLMs to adopt specific social identities can reduce bias \citep{dong2024persona} and mirror human-like ingroup favoritism \citep{hu2025generative}. Incorporating social identity frameworks could enhance user alignment in identity-sensitive contexts \citep{chen-etal-2020-listeners}. Likewise, while bias detection is common, classic \psychTheory{social influence theories} (e.g., conformity, obedience) and \psychTheory{attitude change theories} (e.g., balance theory, cognitive dissonance) are rarely applied to interaction dynamics or bias mitigation, despite their relevance to ethical and socially adaptive behavior. Additionally, malicious actors leveraging social influence can severely undermine trust in digital spaces \citep{ai-etal-2024-defending, liu-etal-2025-propainsight, zeng-etal-2024-johnny}, highlighting the potential of constructs like \psychTheory{inoculation theory} to proactively guard against manipulative strategies.

\textbf{Behavioral psychology} inspires RLHF, yet key concepts like \psychTheory{partial reinforcement}, which improves behavior persistence \citep{ferster1966animal, jensen1961partial}, and \psychTheory{shaping}, which supports gradual learning through successive approximations \citep{LOVE2009421}, are overlooked. Current RLHF relies on uniform rewards, yet behavioral theory warns that flawed rewards can lead to reward hacking. Adding \psychTheory{reward variability} may reduce premature convergence and improve alignment with human intent \citep{amodei2016concreteproblemsaisafety, dayan2008decision}.

\textbf{Personality Psychology} use focuses on \psychTheory{Trait Theory}, overlooking \psychTheory{developmental theories} that explain how individual traits emerge, evolve, and adapt across contexts. These developmental models could enable more coherent and interpretable personality representations, offering a deeper alternative to static prompt-based personas.

\textbf{Cognitive psychology} remains underused, particularly \psychTheory{Schema Theory}, which holds that \psychDef{humans store knowledge as schemas formed through repeated experience} \citep{anderson1984schema}, guiding inference, memory, and learning. Recent work explores schema-inspired methods for compressing user histories and modeling knowledge activation cycles \citep{10849404, Xia02042024}. Further integration may improve long-term context handling and generalization.

%% file: dicussion/discussion_debate.tex
A recurring question is whether human psychology can be directly mapped to LLMs without distortion \citep{lohn-etal-2024-machine-psychology}. Below, we highlight key controversies at this boundary; see Appendix \ref{appendix:debates} for an extended discussion. These challenges motivate new recommendations and highlight open directions for cross-disciplinary exploration.

\paragraph{Terminology Mismatches}
A core tension is the mismatch between psychological terminology and their NLP usage. For example, \textbf{attention} in psychology refers to \textit{selective mental focus}, but in transformers it is a token weighting mechanism without cognitive awareness \citep{lindsay2020attention}, leading to misleading attributions of intentionality. Similarly, \textbf{memory} in psychology entails \textit{structured encoding and recall}, whereas in LLMs it typically refers to context windows or parameters. 

Such anthropomorphic language is increasingly prevalent and shapes public and scholarly assumptions about LLMs, as recent studies show rising human-like descriptors \citep{ibrahim2025thinking}. This calls for disentangling metaphor from mechanism through a precise cross-disciplinary lexicon, preventing both oversimplification and over-anthropomorphization -- an underexplored but crucial research challenge.

\paragraph{Theoretical Discrepancies}
Beyond terminology, deeper theoretical mismatches arise when the NLP community adopts outdated or disputed concepts from psychology. For instance, \psychTheory{predictive coding} \citep{rao1999predictive} is used to analogize LLMs’ next-token prediction, although current research emphasizes hierarchical, multi-scale brain mechanisms \citep{Antonello_Huth, caucheteux2023evidence}. Likewise, folk-psychological typologies like \psychTheory{MBTI} persist in LLM applications despite its criticized validity and reliability \citep{mccrae1989reinterpreting, pittenger1993measuring}. \citet{wagner-etal-2025-mind} positions that \psychTheory{ToM} involves first deciding depth of mentalizing and then applying reasoning accordingly, yet most works focus only on the latter. \psychTheory{Working memory} \citep{baddeley1974working} illustrates another gap: LLM "memory" modules \citep{li-etal-2023-large, kang2024think} do not replicate human constraints, prompting questions about whether AI should emulate human cognitive limits or exceed them for performance gains. \textbf{Behavioral psychology} faces similar critiques \citep{flavell2022emergence, miller2003cognitive}, as RLHF often focuses on reward optimization \citep{ouyang2022training, NEURIPS2023_a85b405e, NEURIPS2024_4147dfaa}, neglecting internal states and risking reward hacking \citep{Krakovna_2020, skalse2022defining}. Broader debates remain over whether LLMs truly “understand” language or function as “stochastic parrots” \citep{AmbridgeBlything20243348, park-etal-2024-multiprageval}.

In response, we recommend refining how psychological theories are mapped into computational models, replacing outdated constructs with supported frameworks, exploring whether human-like constraints aid interpretability, and designing evaluations that track both outputs and internal states. Sustained collaboration between computational and psychological sciences is essential for robust and theory-aligned LLMs.

\paragraph{Evaluation and Validity Debates}
Another major debate is how we evaluate LLM “psychological” abilities -- whether current tests really measure what they claim. For instance, GPT-4 solves around 75\% of false-belief tasks, matching a 6-year-old’s performance \citep{kosinski2024evaluating, strachan2024testing}; some see emergent \psychTheory{ToM}-like reasoning \citep{kosinski2024evaluating}, but others argue it may be pattern matching \citep{strachan2024testing}, noting that minor prompt changes can derail results \citep{shapira-etal-2024-clever}. This calls for more theory-grounded evaluation and clearer definitions.

A parallel controversy involves \textbf{personality}. Some studies find stable simulated personality traits \citep{sorokovikova-etal-2024-llms, huang-etal-2024-reliability}, while others reveal variability under different prompt conditions \citep{gupta-etal-2024-self, shu-etal-2024-dont},  raising questions about inherent
vs. mimicked personas \citet{tseng-etal-2024-two}. 
These debates underscore the need for a systematic, theory-driven framework beyond surface metrics, guiding more faithful replication of human cognition and behavior in LLMs.

These debates also echo concerns raised within \textbf{Psychologists’ Perspectives on LLMs}, where scholars are critically examining the promises and limitations of applying human psychological constructs to LLMs. 
Psychologists are engaging LLMs both as tools for research and as objects of study, but they caution against straightforwardly mapping human psychological constructs onto models trained via next-token prediction on disembodied text. Method papers flag opportunities and pitfalls for using LLMs as participants, judges, or simulators, urging stronger validity checks and domain knowledge when importing instruments or constructs \cite{10.1093/pnasnexus/pgae245}. At the theoretical level, scholars argue we should be precise about what LLMs are meant to model and avoid inferring humanlike learning or cognition from surface similarities \cite{BLANK2023987, 10.1162/opmi_a_00160}. Empirically, comparative studies probe where model behavior diverges from humans on socially loaded tasks: LLMs can pass certain ToM-style probes while still differing on harder variants \cite{strachan2024testing, 10.1162/tacl_a_00674}, and they sometimes amplify cognitive biases such as omission bias in moral judgment \cite{doi:10.1073/pnas.2412015122}. Together, these works show an active psychology literature investigating LLMs’ promise and limits, and underscores why direct transfer of human psychological theory to LLMs requires careful construct validity, task design, and interpretation.

%% file: appendix/psychology_theories.tex
\newcommand{\yes}{\scalebox{1.5}{\textcolor[HTML]{38b000}{\ding{51}}}}
\newcommand{\no}{\scalebox{1.5}{\textcolor[HTML]{d62828}{\ding{55}}}}
\newcommand{\partially}{\scalebox{1.5}{\textcolor[HTML]{ffbe0b}{\ding{117}}}}
\section{Psychology Theories}
\input{graph/psych_theories}

In Tables~\ref{tab:psych_theories_dev_cog_beh}, \ref{tab:psych_theories_soc}, and \ref{tab:psych_theories_per_ling}, we summarize representative psychological theories by sub-area and indicate whether they have been explored in existing LLM research. Table~\ref{tab:psych_theories_dev_cog_beh} covers developmental, cognitive, and behavioral psychology theories; Table~\ref{tab:psych_theories_soc} focuses on social psychology theories; and Table~\ref{tab:psych_theories_per_ling} presents personality psychology and psycholinguistics theories.

For each theory, the “Explored” column captures the extent to which it has been applied in LLM research. The symbol \yes{} denotes multiple surveyed works explicitly leveraging or referencing the theory, \partially{} indicates fewer than three such works, and \no{} signifies that none were identified in our survey. The distribution of these marks highlights which areas of psychological theory have already influenced LLM development—such as working memory theory or reinforcement learning analogies—and which remain largely unexplored, such as social identity theory or certain psycholinguistic processing models.

These tables are designed to provide an at-a-glance view of theoretical coverage and to reveal underexplored opportunities where insights from psychology could inspire new approaches to model different stages of LLMs' development.

%% file: graph/psych_theories.tex
\newcolumntype{C}[1]{>{\centering\arraybackslash}m{#1}}
\newcommand{\smallcite}[1]{\scriptsize{\citep{#1}}}
\newcommand{\gcmidrule}[1]{%
  \arrayrulecolor{lightgray}%
  \cmidrule{#1}%
  \arrayrulecolor{black}%
}

\begin{table*}[h]
\centering
\small

\resizebox{1.0\textwidth}{!}{
\begin{NiceTabular}{C{65pt}|c|C{80pt}|C{250pt}|c}
    \toprule
    \textbf{Psych Area} & \textbf{Sub Area} & \textbf{Theory} & \textbf{Definition} & \textbf{Explored} \\
    \midrule\midrule
    \rowcolor{devcolor}
    \Block{6-1}{\textbf{Developmental Psych}} & \Block{4-1}{Cognitive Development} & \psychTheory{Incremental Cognitive Development} & \psychDef{Children acquire knowledge through sequential tasks with increasing complexity} \smallcite{piaget1976piaget} & \yes \\
    \gcmidrule{3-5}
    \rowcolor{devcolor}
    & & \psychTheory{Scaffolding Theory} & \psychDef{Learning is enhanced through gradually challenging interactions with appropriate guidance} \smallcite{park2009adaptive} & \partially \\
    \gcmidrule{3-5}
    \rowcolor{devcolor}
    & & \psychTheory{Incremental Numerical Understanding} & \psychDef{Numerical concepts are gradually acquired through structured exposure and experience} \smallcite{piaget2013child} & \partially \\
    \gcmidrule{3-5}
    \rowcolor{devcolor}
    & & \psychTheory{Zone of Proximal Development} & \psychDef{Optimal learning occurs in the gap between what a learner can do independently and with assistance} \smallcite{wertsch1988vygotsky} & \partially \\
    \cmidrule{2-5}
    \rowcolor{devcolor}
    & \Block{2-1}{Language Acquisition} & \psychTheory{Language Acquisition Theory} & \psychDef{Language development follows predictable patterns through exposure to linguistic environments} \smallcite{Chomsky_1980} & \partially \\
    \gcmidrule{3-5}
    \rowcolor{devcolor}
    & & \psychTheory{Ecological Validity} & \psychDef{Emphasizes real-world data and environments to mimic natural cognitive development} \smallcite{schmuckler2001ecological} & \partially \\
    \midrule

    \rowcolor{cogcolor}
    \Block{10-1}{\textbf{Cognitive Psych}} & \Block{3-1}{Attention and Perception} & \psychTheory{Selective Attention} & \psychDef{Prioritizes cognitively salient information while filtering out irrelevant stimuli} \smallcite{treisman1969selective_attention} & \yes \\
    \gcmidrule{3-5}
    \rowcolor{cogcolor}
    & & \psychTheory{Top-down and Bottom-up Processing} & \psychDef{Distinguishes between concept-driven (top-down) and data-driven (bottom-up) perceptual processing} \smallcite{gregory_topdown_bottomup} & \yes \\
    \gcmidrule{3-5}
    \rowcolor{cogcolor}
    & & \psychTheory{Predictive Coding} & \psychDef{Anticipatory processing based on prior knowledge and prediction of expected inputs} \smallcite{rao1999predictive} & \partially \\
    \cmidrule{2-5}
    \rowcolor{cogcolor}
    & \Block{3-1}{Memory Systems} & \psychTheory{Working Memory} & \psychDef{Limited-capacity system for temporarily holding and manipulating information} \smallcite{baddeley1974working} & \yes \\
    \gcmidrule{3-5}
    \rowcolor{cogcolor}
    & & \psychTheory{Long-term Memory} & \psychDef{System for storing information over extended periods through semantic organization} \smallcite{tulving1972episodic} & \partially \\
    \gcmidrule{3-5}
    \rowcolor{cogcolor}
    & & \psychTheory{Hippocampal Indexing Theory} & \psychDef{Views the hippocampus as a pointer to neocortical memory representations} \smallcite{teyler1986hippocampal} & \partially \\
    \cmidrule{2-5}
    \rowcolor{cogcolor}
    & \Block{3-1}{Reasoning and Decision Making} & \psychTheory{Cognitive Maturity} & \psychDef{Tthe development and refinement of an individual's thinking, reasoning, and problem-solving abilities} \smallcite{ingersoll1986cognitive} & \yes \\
    \gcmidrule{3-5}
    \rowcolor{cogcolor}
    & & \psychTheory{Theory of Mind} & \psychDef{The ability to attribute mental states to oneself and others and understand others may have different beliefs} \smallcite{baron1985does} & \yes \\
    
    \gcmidrule{3-5}
    \rowcolor{cogcolor}
    & & \psychTheory{Schema Theory} & \psychDef{Knowledge is organized into interconnected patterns that guide processing and interpretation of new information} \smallcite{anderson1984schema} & \no \\

    \midrule
    
    \rowcolor{behavcolor}
    \Block{4-1}{\textbf{Behavioral Psych}} & \Block{4-1}{Learning and Conditioning} & \psychTheory{Classical Conditioning} & \psychDef{Learning occurs when a neutral stimulus becomes associated with a meaningful one}~\smallcite{pavlov2010conditioned} & \partially \\
    \gcmidrule{3-5}
    \rowcolor{behavcolor}
    & & \psychTheory{Operant Conditioning} & \psychDef{Behavior is strengthened or weakened by consequences such as rewards or punishments}~\smallcite{skinner1957behavior,skinner1963operant} & \yes \\
    \gcmidrule{3-5}
    \rowcolor{behavcolor}
    & & \psychTheory{Thorndike’s Law of Effect} & \psychDef{Behaviors followed by satisfying outcomes are more likely to be repeated in the future}~\smallcite{thorndike1927law} & \partially \\
    \gcmidrule{3-5}
    \rowcolor{behavcolor}
    & & \psychTheory{Premack Principle} & \psychDef{A preferred activity can reinforce a less preferred one if access is contingent}~\smallcite{premack1959toward} & \no \\
    \bottomrule

\end{NiceTabular}
}
\captionsetup{font=small}
\caption{Representative developmental, cognitive, and behavioral psychology theories by sub-area. In the “Explored” column, \yes indicates multiple surveyed works, \partially indicates fewer than three, and \no indicates that none emerged in our survey (i.e., not yet substantially explored).}
\label{tab:psych_theories_dev_cog_beh}
\end{table*}

\begin{table*}[h]
\centering
\small

\resizebox{1.0\textwidth}{!}{
\begin{NiceTabular}{C{65pt}|c|C{80pt}|C{250pt}|c}
    \toprule
    \textbf{Psych Area} & \textbf{Sub Area} & \textbf{Theory} & \textbf{Definition} & \textbf{Explored} \\
    \midrule\midrule


    \rowcolor{soccolor}
    \Block{18-1}{\textbf{Social Psych}} & \Block{3-1}{Social Cognition} & \psychTheory{Attribution Theory} & \psychDef{Explains how people infer causes of behavior as internal or external} \smallcite{fiske2020social, baron2012science} & \no \\
    \gcmidrule{3-5}
    \rowcolor{soccolor}
    & & \psychTheory{Dual-Process Theory} & \psychDef{Differentiates between fast, intuitive (System 1) and slow, deliberate (System 2) reasoning} \smallcite{kahneman2011thinking} & \yes \\
    \gcmidrule{3-5}
    \rowcolor{soccolor}
    & & \psychTheory{Theory of Mind (ToM)} & \psychDef{How individuals understand and attribute mental states to others} \smallcite{baron1985does} & \yes \\
    \cmidrule{2-5}
    
    \rowcolor{soccolor}
    & \Block{4-1}{Social Influence} & \psychTheory{Social Impact Theory} & \psychDef{The magnitude of social influence depends on the strength, immediacy, and number of sources} \smallcite{latane1981psychology} & \no \\
    \gcmidrule{3-5}
    \rowcolor{soccolor}
    & & \psychTheory{Conformity Theories} & \psychDef{Explore how group pressure can alter individual judgments} \smallcite{asch2016effects} & \yes \\
    \gcmidrule{3-5}
    \rowcolor{soccolor}
    & & \psychTheory{Obedience Theories} & \psychDef{Demonstrate how authority influences behavior, highlighting conditions under which individuals comply} \smallcite{milgram1963behavioral} & \no \\
    \gcmidrule{3-5}
    \rowcolor{soccolor}
    & & \psychTheory{Persuasion Models} & \psychDef{Explain how messages processed via central or peripheral routes can lead to attitude change} \smallcite{petty2012communication} & \yes \\
    \cmidrule{2-5}

    \rowcolor{soccolor}
    & \Block{2-1}{Group Dynamics} & \psychTheory{Groupthink} & \psychDef{Examines how the desire for conformity and group cohesion can lead to flawed decision-making and suppression of dissenting opinions} \smallcite{janis1972victims} & \no \\
    \gcmidrule{3-5}
    \rowcolor{soccolor}
    & & \psychTheory{Social Facilitation and Social Loafing} & \psychDef{Investigates how the presence of others can enhance performance on simple tasks or reduce effort in collective work} \smallcite{zajonc1965social, latane1979many} & \no \\
    \cmidrule{2-5}

    \rowcolor{soccolor}
    & \Block{4-1}{Attitude Change} & \psychTheory{Cognitive Dissonance Theory} & \psychDef{Explains how inconsistencies between beliefs or behaviors create discomfort, prompting attitude change to restore consistency} \smallcite{morvan2017analysis} & \partially \\
    \gcmidrule{3-5}
    \rowcolor{soccolor}
    & & \psychTheory{Elaboration Likelihood Model (ELM)} & \psychDef{Proposes that persuasion occurs via a central route (deliberate processing) or a peripheral route (heuristic processing), depending on the recipient’s motivation and capacity} \smallcite{petty2012communication} & \no \\
    \gcmidrule{3-5}
    \rowcolor{soccolor}
    & & \psychTheory{Balance Theory} & \psychDef{Suggests that individuals strive for consistency among their attitudes and relationships, adjusting beliefs to maintain cognitive harmony} \smallcite{heider1946attitudes} & \no \\
    \gcmidrule{3-5}
    \rowcolor{soccolor}
    & & \psychTheory{Inoculation Theory} & \psychDef{Posits that exposure to weak counterarguments can strengthen resistance to persuasion by preemptively activating defensive mechanisms} \smallcite{mcguire1964inducing} & \no \\
    \cmidrule{2-5}

    \rowcolor{soccolor}
    & \Block{5-1}{Self and Identity} & \psychTheory{Self-Reflection} & \psychDef{Defines the process of introspection, with attention placed on the self-concept} \smallcite{Phillips2020} & \yes \\
    \gcmidrule{3-5}
    \rowcolor{soccolor}
    & & \psychTheory{Self-Perception Theory} & \psychDef{Explains how individuals infer their internal states by observing their own behavior} \smallcite{bem1972self} & \no \\
    \gcmidrule{3-5}
    \rowcolor{soccolor}
    & & \psychTheory{Social Identity Theory} & \psychDef{Posits that group membership shapes self-concept and influences intergroup behavior} \smallcite{tajfel1979integrative} & \partially \\
    \gcmidrule{3-5}
    \rowcolor{soccolor}
    & & \psychTheory{Self-Categorization Theory} & \psychDef{Expands on social identity theory, describing how individuals classify themselves and others into social groups, shaping social norms} \smallcite{maines1989rediscovering} & \no \\
    \gcmidrule{3-5}
    \rowcolor{soccolor}
    & & \psychTheory{Self-Affirmation Theory} & \psychDef{Suggests that individuals are motivated to maintain their self-integrity when faced with threats to their self-concept} \smallcite{steele1988psychology} & \no \\
    \bottomrule
\end{NiceTabular}
}
\captionsetup{font=small}
\caption{Representative social psychology theories by sub-area. In the “Explored” column, \yes indicates multiple surveyed works, \partially indicates fewer than three, and \no indicates that none emerged in our survey (i.e., not yet substantially explored).}
\label{tab:psych_theories_soc}
\end{table*}

\begin{table*}[h]
\centering
\small

\resizebox{1.0\textwidth}{!}{
\begin{NiceTabular}{C{65pt}|c|C{80pt}|C{250pt}|c}
    \toprule
    \textbf{Psych Area} & \textbf{Sub Area} & \textbf{Theory} & \textbf{Definition} & \textbf{Explored} \\
    \midrule\midrule
    
    \rowcolor{perscolor}
    \Block{8-1}{\textbf{Personality Psych}} & \Block{3-1}{Personality traits} & \psychTheory{Big Five Model} & \psychDef{The Five-Factor Model (FFM), also known as OCEAN, categorizes personality into five dimensions: Openness to experience, Conscientiousness, Extraversion, Agreeableness, Neuroticism} \smallcite{doi:10.1177/0146167202289008} & \yes \\
    \gcmidrule{3-5}
    \rowcolor{perscolor}
    & & \psychTheory{Myers-Briggs Type Indicator (MBTI)} & \psychDef{Classifies individuals into 16 personality types based on four dichotomies (e.g., Introversion vs. Extraversion)  \smallcite{myers1995gifts}. While widely used, MBTI has been criticized for lacking empirical validity, reliability, and independence between its categories. } \smallcite{pittenger1993measuring} & \yes \\
    \gcmidrule{3-5}
    \rowcolor{perscolor}
    & & \psychTheory{Eysenck Personality Questionnaire-Revised (EPQR-A)} & \psychDef{Contains a 24-item personality test that measures extraversion, neuroticism, psychoticism, and social desirability.} \smallcite{eysenck1984eysenck} & \yes \\
    \cmidrule{2-5}

    \rowcolor{perscolor}
    & \Block{5-1}{Personality Theories} & \psychTheory{Humanistic Theory} & \psychDef{ Emphasizes free will, personal growth, and self-actualization. This perspective focuses on individuals’ subjective experiences and their drive to achieve their full potential.} \smallcite{stefaroi2015humanistic} & \no \\
    \gcmidrule{3-5}
    \rowcolor{perscolor}
    & & \psychTheory{Psychoanalytic Theory} & \psychDef{Originating from Freud, this theory conceptualizes personality as the dynamic interplay between the id, ego, and superego, with unconscious processes playing a central role in shaping behavior.} \smallcite{scharff2013psychoanalytic} & \no \\
    \gcmidrule{3-5}
    \rowcolor{perscolor}
    & & \psychTheory{Behaviorist Theory} & \psychDef{Views personality as a set of learned responses shaped by environmental reinforcements and punishments. This perspective, pioneered by figures like Skinner and Watson, rejects internal mental states in favor of observable behaviors.} \smallcite{pierce2008behavior} & \no \\
    \gcmidrule{3-5}
    \rowcolor{perscolor}
    & & \psychTheory{Social Cognitive Theory} & \psychDef{Highlights the role of cognitive processes in personality, emphasizing how expectations, beliefs, and observational learning shape behavior.} \smallcite{spielman2024psychology} & \no \\
    \gcmidrule{3-5}
    \rowcolor{perscolor}
    & & \psychTheory{Trait Theory} & \psychDef{Focuses on identifying and measuring stable personality traits that influence behavior across different contexts.} \smallcite{cartwright1979theories} & \partially \\
    \midrule

    \rowcolor{lingcolor}
    \Block{8-1}{\textbf{Psycholinguistics}} & \Block{2-1}{Language Acquisition} & \psychTheory{Universal Grammar} & \psychDef{Proposes an innate linguistic capacity that guides language learning} \smallcite{chomsky1957syntactic, chomsky1965aspects}&\yes \\
    \gcmidrule{3-5}
    \rowcolor{lingcolor}
    & & \psychTheory{Usage-Based Theory}              & \psychDef{Emphasizes the role of social interaction and cognitive processes in language learning, rather than innate universal grammatical structures} \smallcite{tomasello2005constructing} & \no \\ 
    \cmidrule{2-5}
    \rowcolor{lingcolor}
    & \Block{4-1}{Language Comprehension}& \psychTheory{Garden Path Theory}             & \psychDef{Describes how people backtrack and reanalyze the sentence structure when encountering unexpected linguistic elements that challenge their initial understanding} \smallcite{FRAZIER1982178} & \partially \\
    \gcmidrule{3-5}
    \rowcolor{lingcolor}
    & & \psychTheory{Constraint-Based Models}         & \psychDef{Language processing is an interactive, probabilistic process where multiple sources of information simultaneously contribute to understanding, rather than following a strict, sequential parsing approach} \smallcite{macdonald1994lexical}& \partially\\
    \gcmidrule{3-5}
    \rowcolor{lingcolor}
    & & \psychTheory{Good-Enough Processing}          & \psychDef{Proposes that humans comprehend language through approximate, semantically-focused representations that capture the core meaning rather than constructing syntactically perfect linguistic interpretations} \smallcite{ferreira2007good} & \no \\
    \gcmidrule{3-5}
    \rowcolor{lingcolor}
    & & \psychTheory{Construction-Integration Model}  & \psychDef{Describes text comprehension as a two-stage process where readers first generate multiple, loosely connected propositions and then systematically filter and integrate them into a coherent, meaningful understanding.} \smallcite{kintsch1988role}& \no \\
    \cmidrule{2-5}
    \rowcolor{lingcolor}
    & \Block{2-1}{Language Production} & \psychTheory{WEAVER++ Model}                  & \psychDef{Comprehensive framework for speech production as a complex, multi-stag, parallel process} \smallcite{levelt1999theory} & \no \\
    \gcmidrule{3-5}
    \rowcolor{lingcolor}
    & & \psychTheory{Interactive Two-Step Model}      & \psychDef{An interactive, probabilistic process of lexical selection and phonological encoding, where multiple linguistic levels simultaneously influence each other during speech generation} \smallcite{goldrick2007lexical} & \no\\
    \bottomrule
\end{NiceTabular}
}
\captionsetup{font=small}
\caption{Representative personality psychology and psycholinguistics theories by sub-area. In the “Explored” column, \yes indicates multiple surveyed works, \partially indicates fewer than three, and \no indicates that none emerged in our survey (i.e., not yet substantially explored).}
\label{tab:psych_theories_per_ling}
\end{table*}

%% file: appendix/keywords_validation.tex
\section{Search Strategy and Keyword Lists}
\label{appendix:keywords}

\subsection{Search Strategy and Validation}
We survey 227 papers from major *CL venues, plus COLING, NeurIPS, ICML, ICLR, and influential arXiv preprints, from 2021 to 2025.

\subsubsection{Search strategy:}

       Each author was assigned specific psychological domains (domains were consulted with psychology experts to ensure no major areas were overlooked).
       Each paper list was cross-checked by other authors.
       
      Full keywords combined psychological terms (e.g., “working memory,” “theory of mind,” “operant conditioning”) with LLM-related terms (e.g., “language model,” “transformer”) in systematic combinations. Full keyword list is provided below.
       When in doubt, cross-verification was conducted with both psychology and NLP experts

\subsubsection{Validity of connections:}
    All psychological connections were rigorously validated through a multi-step process:
    \begin{enumerate}
      \item Initial connections identified and confirmed by our team of 5 NLP experts, one of which with a degree in psychology, ensuring both technical and theoretical grounding
      \item Cross-verification conducted across the entire team, with consultation of external psychology experts when connections required specialized domain knowledge
      \item Final systematic review by senior co-authors in both NLP \& Psychology, 2 psychologists with expertise spanning both psychology research and NLP applications
    \end{enumerate}
    This multi-layered validation process ensures that every psychological theory-LLM connection in our survey is both theoretically sound and technically feasible.

\subsection{Keyword Lists}
\subsubsection{Developmental Psychology}
\textbf{Subareas:} cognitive development; language acquisition (merged into psycholinguistics)\\
\textbf{Keywords:} “piaget”, “cognitive development”, “vygotsky”, “sociocultural development”, “scaffolding”, “social learning”, “zone of proximal”, “observational learning”, “moral development”, “ecological validity”, “ecological systems”, “constructivist”, “constructive development” \\
\textbf{Reference table:} \textbf{Reference table:} Table~\ref{tab:psych_theories_dev_cog_beh}

\subsubsection{Cognitive Psychology}
\textbf{Subareas:} Perception; Attention; Memory; Reasoning \& Decision Making \\
\textbf{Keywords:} “perception”, “top down”, “bottom up”, “contextual information”, “schema theory”, “schemas”, “pattern recognition”, “constructivist”, “knowledge construction”, “predictive coding”, “attention psychology”, “selective attention”, “memory psychology”, “working memory”, “memory augmentation”, “long-term memory”, “knowledge retention”, “episodic memory”, “hippocampal indexing”, “cognitive load”, “dual-process”, “cognitive maturity”, “cognitive biases”, “metacognition”, “metacognitive learning”, “self-reflection”, “theory of mind” (the keyword “psychology” was appended during search as well) \\
\textbf{Reference table:} Table~\ref{tab:psych_theories_dev_cog_beh}

\subsubsection{Behavioral Psychology}
\textbf{Subareas:} Classical Conditioning; Operant Conditioning; Observational Learning (Social Learning); Behavior Modification and Applied Behavior Analysis \\
\textbf{Keywords:} ``Behavioral psychology'', ``behaviorism'', ``classical conditioning psychology'', ``Pavlovian conditioning'', ``unconditioned stimulus'', ``unconditioned response'', ``conditioned stimulus'', ``conditioned response'', ``neutral stimulus'', ``acquisition learning'', ``extinction'', ``spontaneous recovery'', ``stimulus generalization'', ``stimulus discrimination'', ``higher-order conditioning'', ``second-order conditioning'', ``operant conditioning'', ``RLHF'', ``RLAIF'', ``instrumental conditioning'', ``law of effect'', ``reinforcement learning'', ``reward'', ``positive reinforcement'', ``negative reinforcement'', ``punishment'', ``positive punishment'', ``negative punishment'', ``discriminative stimulus'', ``shaping'', ``chaining'', ``primary reinforcer'', ``secondary reinforcer'', ``conditioned reinforcer'', ``continuous reinforcement'', ``partial reinforcement'', ``intermittent reinforcement'', ``fixed interval schedule'', ``variable interval schedule'', ``fixed ratio schedule'', ``variable ratio schedule'', ``observational learning'', ``modeling psychology'', ``imitation'', ``vicarious reinforcement'', ``vicarious punishment'', ``behavior modification'', ``behavior therapy'', ``Applied Behavior Analysis'', ``token economy'', ``aversion therapy'', ``aversive conditioning'', ``contingency management'' \\
\textbf{Reference table:} Table~\ref{tab:psych_theories_dev_cog_beh}

\subsubsection{Social Psychology}
\textbf{Subareas:} social cognition; social influence; group dynamics; attitude change; self \& identity \\
\textbf{Keywords:} social cognition; social influence; group dynamics; attitude change; self and identity; attribution theory; dual-process; theory of mind; social impact; conformity; obedience; persuasion; groupthink; social facilitation; social loafing; cognitive dissonance; elaboration likelihood model; balance theory; inoculation theory; self-reflection; self-perception; social identity; self-categorization; self-affirmation \\
\textbf{Reference table:} Table~\ref{tab:psych_theories_soc}

\subsubsection{Personality Psychology}
\textbf{Subareas:} humanistic theory; psychoanalytic theory; behaviorist theory; social cognitive theory; trait theory (used in combination with ``personality'') \\
\textbf{Keywords:} ``personality'', ``personality psychology'', ``personality traits'', ``the Big Five'', ``Big Five Model'', ``OCEAN'', ``Myers-Briggs Type Indicator'', ``MBTI'', ``EPQR-A'', ``Eysenck Personality Questionnaire'', ``Socionics'', ``temperaments'', ``Personality Factors'' \\
\textbf{Reference table:} Table~\ref{tab:psych_theories_per_ling}

\subsubsection{Psycholinguistics}
\textbf{Keywords:} psycholinguistic; linguistic; phonology/phonological; phonetic; morphology/morphological; semantic; syntax/syntactic; pragmatic \\
\textbf{Reference table:} Table~\ref{tab:psych_theories_per_ling}

%% file: appendix/rlhf.tex
\section{Extended Discussion on Reinforcement Learning from Human Feedback (RLHF)}
\label{appendix:rlhf}

\subsection{Operant Conditioning in RLHF}
During RLHF fine-tuning, the model (agent) generates responses while a learned reward function $R(x)$, often a neural network trained on preference data, assigns scores to candidate outputs $x$. These scores proxy for human judgment and guide policy updates to reinforce higher-reward behaviors. For instance, \cite{ouyang2022training} trains InstructGPT via Proximal Policy Optimization \citep{schulman2017proximal}: responses deemed more helpful or accurate by human evaluators receive greater reward, whereas undesirable or incorrect outputs face penalization. Unlike purely exploration-based RL methods, this arrangement leverages human insight to provide a more precise learning signal; however, success relies on careful and consistent reward modeling that captures subtle human values.

\subsection{Modeling Human Preferences as a Reward Function}
Although extensive work has been conducted in RLHF, here we primarily highlight recent approaches or methodologies explicitly grounded in psychological theories. Building robust reward functions from heterogeneous or ambiguous feedback remains a core challenge in RLHF. Early foundational frameworks \citep{NIPS2017_d5e2c0ad, stiennon2022learningsummarizehumanfeedback} laid essential groundwork for converting human judgments into usable reward signals, drawing implicitly from principles of \psychTheory{Operant Conditioning Theory}. More recent advancements explicitly target improvements in stability, scalability, and fairness, addressing issues arising from the inherent variability and complexity of human preferences.

\cite{NEURIPS2023_a85b405e} introduced Direct Preference Optimization (DPO), simplifying preference integration by directly optimizing the policy through a closed-form solution, thus removing the need for explicit intermediate reward modeling. Extending these efforts toward equitable alignment, \cite{NEURIPS2024_4147dfaa} proposed Group Robust Preference Optimization (GRPO), ensuring robustly aligned outcomes across diverse demographic groups, addressing biases commonly observed in human-driven reward processes.

Further refinements emphasize enhancing alignment accuracy through psychological considerations. For instance, Contrastive Preference Learning \citep{hejna2024contrastive} utilizes regret-based losses inspired by behavioral economics, facilitating stable off-policy learning without conventional RL techniques. Distributional Preference Learning \citep{siththaranjan2024distributional} aligns reward modeling more closely with human cognitive patterns by capturing human values as probability distributions rather than point estimates. Variational Preference Learning (VPL) \citep{poddar2024personalizing} further integrates psychological realism, introducing latent-variable modeling to personalize RLHF, reflecting variability in individual user preferences rather than imposing a universal reward structure.

These advancements collectively illustrate how psychological theory, particularly \psychTheory{Operant Conditioning Theory}, continues to shape and inspire sophisticated techniques for reliably aligning LLM behavior with nuanced human values.

\subsection{Reinforcement Schedules and Feedback Frequency}
In early RLHF, feedback is typically sparse — a single scalar reward per output — which causes a credit assignment problem: the model can't tell which parts of the output led to the reward. This is similar to delayed feedback in animal learning, which slows progress.
Psychology shows that immediate and frequent reinforcement improves learning. Similarly, recent RLHF methods provide dense, token-level feedback (e.g., from a critic model), which improves sample efficiency and training stability. To address this, \cite{cao-etal-2024-enhancing} propose LLM self-critique, a method that uses a secondary model to provide dense, token-level feedback during generation. This simulates a continuous reinforcement schedule, analogous to real-time feedback in behavioral training, and leads to more stable and efficient learning.
Another factor is how often feedback is given: continuous vs. partial reinforcement. While human feedback is often sparse due to cost, using AI feedback models (like RLAIF, will discuss later) allows for more frequent feedback. Even with limited human scores, techniques like credit assignment can distribute reward across the output.

\subsection{Reward Prediction Errors as a Learning Driver}

At the heart of reinforcement learning lies the concept of reward prediction error (RPE), which arises when there is a discrepancy between an agent's expected reward and the reward it actually receives, prompting adjustments and driving learning \citep{sutton2018reinforcement}. This mechanism closely parallels dopaminergic signaling in animal brains, where dopamine neurons respond strongly to unexpected rewards or punishments, effectively reinforcing behaviors associated with positive surprises or reducing those linked to disappointments \citep{schultz1998predictive}. In RLHF, reward prediction errors similarly guide model updates; each model output receives a score from a reward model trained on human preferences, and deviations between these scores and the model’s predicted rewards are used to adjust behavior. However, simplistic or flawed reward models can lead to "reward hacking," where the model exploits blind spots in the reward function rather than genuinely aligning with human values \citep{amodei2016concreteproblemsaisafety}. Introducing variability in reward signals can encourage exploration and mitigate premature convergence on suboptimal strategies \citep{dayan2008decision}. To address reward hacking and reward-model inconsistencies, recent approaches have formulated RLHF as a constrained Markov decision process with dynamic weighting \citep{moskovitz2024confronting}, introduced information-theoretic regularization techniques (InfoRM) \citep{miao2024inform}, and proposed methods such as ConvexDA and reward fusion to stabilize and enhance reward-model consistency \citep{shen2024the}.

\subsection{Implications for Bias, Alignment, and Reward Modeling}
Employing these behavioral principles may improve how well RLHF handles biases and achieves robust alignment. For instance, diverse trainers and variable scenarios can prevent conditioning bias, where the model overfits to a narrow segment of human preferences \citep{sheng-etal-2019-woman}. Moreover, shaping and multi-dimensional reward functions can address multiple alignment goals simultaneously (e.g., factual accuracy and polite style), limiting reward hacking. 

At the same time, grounding RLHF in behavioral theory highlights persistent pitfalls. Models still lack an intrinsic understanding of human values, and an imprecise reward signal can reinforce superficial behaviors. To mitigate these risks, a cycle of model auditing, reward model refinement, and re-training can mirror how animal trainers continually adjust reinforcement to avoid unwanted side effects. 

%% file: appendix/dialgue.tex
\section{Persona-Inspired Dialogue Generation}
\label{appendix:persona_dialogue}
Personality has also inspired improvements truthfulness, response grounding, and broader alignments. \citet{zhang-etal-2024-self-contrast} introduced Self-Contrast to enhance internal consistency, and \citet{joshi-etal-2024-personas} proposed the Persona Hypothesis, linking truthfulness to pretraining structure.  \citet{kim-etal-2024-panda} introduced PANDA to reduce persona overuse in dialogue. \citet{zhang-etal-2024-self-contrast} introduced a reflection-based technique to reduce internal inconsistencies. \citet{lee-etal-2025-spectrum} models multidimensional self-concept to enhance authenticity. \citet{joshi-etal-2024-personas} proposed the Persona Hypothesis, arguing that LLMs encode truthful and untruthful personas from their training distribution. \citet{kim-etal-2024-panda} addressed the overuse of persona cues to improve contextual appropriateness. Persona-guided generation has been applied to emotionally supportive role-play settings \citep{ye-etal-2025-sweetiechat, chen-etal-2025-comif}.

%% file: appendix/debates.tex
\section{Extended Discussion on Debates over NLP-Psychology Intersection}
\label{appendix:debates}

A recurring theme is whether human psychology can be naively mapped onto LLM behavior without distortion \citep{lohn-etal-2024-machine-psychology}. Therefore, in this section, we discuss several major points of contention at this interdisciplinary boundary. These issues motivate a set of recommendations and highlight open directions for future cross-disciplinary research.

\paragraph{Terminology Mismatches}
One key issue is the mismatch in terminology and the anthropomorphization of technical concepts. Terms like \psychTheory{attention}, \psychTheory{memory}, and “understanding” have specific meanings in psychology that differ from their usage in NLP. For instance, \textbf{attention} in psychology refers to \psychTheory{selective mental focus and executive control}, whereas in transformers models, it is a mathematical mechanism for weighting tokens -- without cognitive awareness \citep{lindsay2020attention}. This divergence can lead to misleading interpretations, such as assuming models exhibit intentional focus when they merely perform matrix operations. Similar misalignments exist for terms like \textbf{memory} (which in psychology implies \psychTheory{a structured encoding and recall process}, versus an LLM’s context window or weight parameters) and expressions such as “knows” or “thinks.”

Such anthropomorphic language is increasingly prevalent and shapes public and scholarly assumptions about LLMs. Recent analyses have found a growing prevalence of human-like descriptors for LLM behavior, raising calls to carefully disentangle metaphor from mechanism  \citep{ibrahim2025thinking}. An open research direction is developing a more precise cross-disciplinary lexicon: how can we describe model behaviors in ways that neither oversimplify the psychology nor over-anthropomorphize the engineering? Improving interdisciplinary communication by explicitly defining terms and drawing careful analogies remains an important but under-addressed challenge.

\paragraph{Theoretical Discrepancies in Use of Psychology}
Beyond terminology, discrepancies arise in the adoption of psychological theories within NLP research. Sometimes, NLP integrates concepts from psychology that are outdated or contested in their original fields.  For instance, \psychTheory{predictive coding}, which proposes that the brain continuously anticipates sensory input and updates via prediction errors \citep{rao1999predictive}, is often used as a metaphor for LLMs’ next-token prediction. However, contemporary studies emphasize that brain prediction operates across hierarchical and multi-scale structures \citep{Antonello_Huth, caucheteux2023evidence}, cautioning against simplistic analogies that risk misrepresenting the theory.

Another example is the lingering use of folk-psychological typologies like the \psychTheory{MBTI} in some LLM studies. Despite its cultural popularity, \psychTheory{MBTI} has faced substantial criticism for poor validity and reliability \citep{pittenger1993measuring}. It classifies personality into 16 types based on Jungian dichotomies; however, research indicates these categories lack stability and predictive power regarding behavior \citep{mccrae1989reinterpreting}. Nonetheless, the ease of obtaining of MBTI-labeled data has led some NLP studies to treat these categories as definitive, highlighting a theoretical lag where NLP adopts psychological models that mainstream psychology has largely moved beyond.

\psychTheory{Working memory} presents another gap. While cognitive psychology and neuroscience characterize it by limited capacity and active attention control \citep{baddeley1974working}, LLM approximations -- such as short-term retention modules \citep{kang2024think} or memory mechanisms for external context \citep{li-etal-2023-large} -- do not replicate these constraints. This raises questions: Should AI systems emulate human cognitive limitations to achieve more human-like reasoning, or should they leverage their capacity to surpass such constraints? If certain human limitations, like bounded memory, lead to desirable properties such as better interpretability or reduced distractions, might it be useful to impose similar limits on AI? These questions remain largely open.

Finally, a related debate concerns behavioral psychology. The field has been critiqued for ignoring cognitive processes \citep{miller2003cognitive} and internal mental states \citep{flavell2022emergence} that drive the observed behaviors, limiting its explanatory power. With the critiques remaining, the superficial application of behavioral psychology is also evident in LLM research. For instance, RLHF draws from \psychTheory{operant conditioning} but largely focuses on optimizing rewards \citep{ouyang2022training, NEURIPS2023_a85b405e, NEURIPS2024_4147dfaa}, often neglecting internal model states. Consequently, a flip-side of such optimization is reward hacking \citep{skalse2022defining}, where models exploit shortcuts without meeting true objectives -- mirroring human behavior under evaluative pressure \citep{Krakovna_2020}. Deeper integration of cognitive psychology is needed to address these limitations in LLM design.

The debate over whether LLMs possess a true understanding of language or merely function as "stochastic parrots" \citep{10.1145/3442188.3445922} remains ongoing. Linguists have largely been skeptical \citep{AmbridgeBlything20243348}, arguing that language ability is inherently abstract and complex, extending beyond mere statistical pattern recognition. \cite{park-etal-2024-multiprageval} connection between mathematical reasoning and high-level linguistic comprehension.

\paragraph{Evaluation and Validity Debates}
Anoter central debate concerns how we evaluate LLMs on purportedly “psychological” abilities -- and whether current tests measure what we assume. For example, advanced LLMs like GPT-4 perform well on traditional \psychTheory{ToM} tasks, solving around 75\% of false-belief scenarios, comparable to a 6-year-old child \citep{kosinski2024evaluating, strachan2024testing}. Some interpret this as emergent ToM-like reasoning \citep{kosinski2024evaluating}, but others caution that high performance may reflect surface-level pattern matching rather than genuine mental-state attribution. Researchers emphasize that correct answers do not imply mentalizing ability \citep{strachan2024testing}, and minor prompt changes can significantly impair model performance \citep{shapira-etal-2024-clever}. This underscores the need for more rigorous, theory-grounded evaluations and clearer cross-disciplinary definitions.

A similar controversy surrounds personality modeling. Some studies suggest LLMs exhibit stable simulated personality traits \citep{sorokovikova-etal-2024-llms, huang-etal-2024-reliability}, enabling consistent persona simulation across prompts. However, others show that LLM responses vary with prompt framing and response order, undermining test reliability \citep{gupta-etal-2024-self, shu-etal-2024-dont}. \citet{tseng-etal-2024-two} distinguish between role-playing (adopting assigned traits) and personalization (adapting to users), raising a fundamental question: do LLMs have inherent personalities, or merely mimic behavior? While LLMs can simulate personality, inconsistent assessments cast doubt on whether such traits are emergent or engineered -- an open direction for future work.

In summary, these debates highlight the need for a systematic, theory-driven framework that goes beyond superficial performance metrics, thereby enhancing model interpretability and guiding the development of LLMs to more faithfully replicate the complexities of human cognition and behavior.